\documentclass{ieeeaccess}
\usepackage{cite}
\usepackage{amsmath,amssymb,amsfonts}
\usepackage{algorithmic}
\usepackage{graphicx}
\usepackage{textcomp}
\usepackage{url}
\usepackage{array}

\usepackage{bm}
\usepackage{etoolbox}
\makeatletter
\AtBeginDocument{\DeclareMathVersion{bold}
\SetSymbolFont{operators}{bold}{T1}{times}{b}{n}
\SetSymbolFont{NewLetters}{bold}{T1}{times}{b}{it}
\SetMathAlphabet{\mathrm}{bold}{T1}{times}{b}{n}
\SetMathAlphabet{\mathit}{bold}{T1}{times}{b}{it}
\SetMathAlphabet{\mathbf}{bold}{T1}{times}{b}{n}
\SetMathAlphabet{\mathtt}{bold}{OT1}{pcr}{b}{n}
\SetSymbolFont{symbols}{bold}{OMS}{cmsy}{b}{n}
\renewcommand\boldmath{\@nomath\boldmath\mathversion{bold}}}
\makeatother

\newcommand{\prompt}[1]{\emph{#1}}

\DeclareFontShape{T1}{formata}{m}{sl}{<->ssub * formata/m/it}{}

\makeatletter
\def\subsection{\@startsection{subsection}{2}{1pt}{-1pc plus -1pt minus -1pt}%
              {1pt}{\noindent\raggedright\color{greycolor}\sectionBfont\uppercase}}%
\def\subsubsection{\@startsection{subsubsection}{3}{1pt}{-1pc plus -1pt minus -1pt}%
              {1pt}{\raggedright\sectionCfont\hspace*{-1pt}\color{greycolor}}}%
\makeatother

\makeatletter
\def\@dropvolume#1{\patchcmd#1{{\footervolfont VOLUME\ \thevol, \theyear}}{}{}%
  {\errmessage{Failed to remove the VOLUME footer}}}
\@dropvolume\ps@titlepage \@dropvolume\ps@titlepage
\@dropvolume\ps@headings \@dropvolume\ps@headings
\makeatother

\usepackage[colorlinks,allcolors=blue]{hyperref}

\begin{document}
\bstctlcite{CTL:access}
\history{This work has been submitted to the IEEE for possible publication.
Copyright may be transferred without notice, after which this version
may no longer be accessible.}
\makeatletter
\patchcmd{\@maketitle}{{\vss\doifont Digital Object Identifier\space\@doi\vss}}{\vspace*{-1.2pt}}{}
  {\errmessage{Failed to remove the Digital Object Identifier line}}
\makeatother

\title{CCG Diffusion: \\
Combinatory Categorial Grammar--Driven \\
Compositional and Syntactically Disambiguated \\
Text-to-Image Generation}
\author{\uppercase{Ryosuke Yamaki}\authorrefmark{1},
\uppercase{Daichi Mochihashi}\authorrefmark{2},
\uppercase{Nobutaka Shimada}\authorrefmark{1}, \IEEEmembership{Member, IEEE},
\uppercase{Seitaro~Shinagawa}\authorrefmark{3}, AND
\uppercase{Tadahiro Taniguchi}\authorrefmark{4,1}, \IEEEmembership{Member, IEEE}}

\address[1]{Ritsumeikan University, Ibaraki, Osaka 567-8570, Japan (e-mail: yamaki.ryosuke@em.ci.ritsumei.ac.jp; shimada@ci.ritsumei.ac.jp)}
\address[2]{The Institute of Statistical Mathematics, Tachikawa, Tokyo 190-8562, Japan (e-mail: daichi@ism.ac.jp)}
\address[3]{Nara Institute of Science and Technology, Ikoma, Nara 630-0192, Japan (e-mail: sei.shinagawa@is.naist.jp)}
\address[4]{Kyoto University, Sakyo-ku, Kyoto 606-8501, Japan (e-mail: taniguchi@i.kyoto-u.ac.jp)}
\tfootnote{This work was supported by JST ACT-X, Japan, Grant Number JPMJAX24CS.}

\markboth
{Yamaki \headeretal: CCG Diffusion}
{Yamaki \headeretal: CCG Diffusion}

\corresp{Corresponding author: Ryosuke Yamaki (e-mail: yamaki.ryosuke@em.ci.ritsumei.ac.jp).}

\begin{abstract}
Text-to-image generation models do not explicitly receive the syntactic structure of the
prompt, i.e., which word modifies which and where each phrase attaches.
Consequently, they often struggle to bind attributes to the correct objects, to
depict spatial relations and counts correctly, and to render syntactically
ambiguous prompts in the intended interpretation. Existing syntax-based
control methods mainly address attribute binding and have limited ability to
control spatial relations and counts or to specify the interpretation of
syntactically ambiguous prompts. They rely on phrase-structure or dependency
grammars and require additional rules or constraints on the tree structure to
extract modification relations between words or to specify a syntactic
interpretation. We propose
CCG Diffusion, a training-free method that controls image generation using
Combinatory Categorial Grammar (CCG). In CCG, the category of each word
carries rich syntactic information: it specifies what the word combines with
and what phrase it forms. Modification relations between words can therefore
be read directly from a CCG parse. Moreover, prepositional phrase (PP)
attachment ambiguity, a representative syntactic ambiguity, can be
resolved simply by fixing the preposition's category. CCG Diffusion
applies the same two CCG-based mechanisms, prompt rewriting and attention
guidance, both to controlling attribute binding, spatial relations, and counts
and to resolving PP ambiguity. On two architecturally different backbones,
Stable Diffusion and Sana, CCG Diffusion achieves the highest scores among the
compared methods in attribute binding and 2D spatial relations and
significantly outperforms all of them in PP ambiguity resolution.
\end{abstract}


\titlepgskip=-21pt

\maketitle

\section{Introduction}
\label{sec:introduction}
\begin{figure*}[t]
\centering
\includegraphics[width=\textwidth]{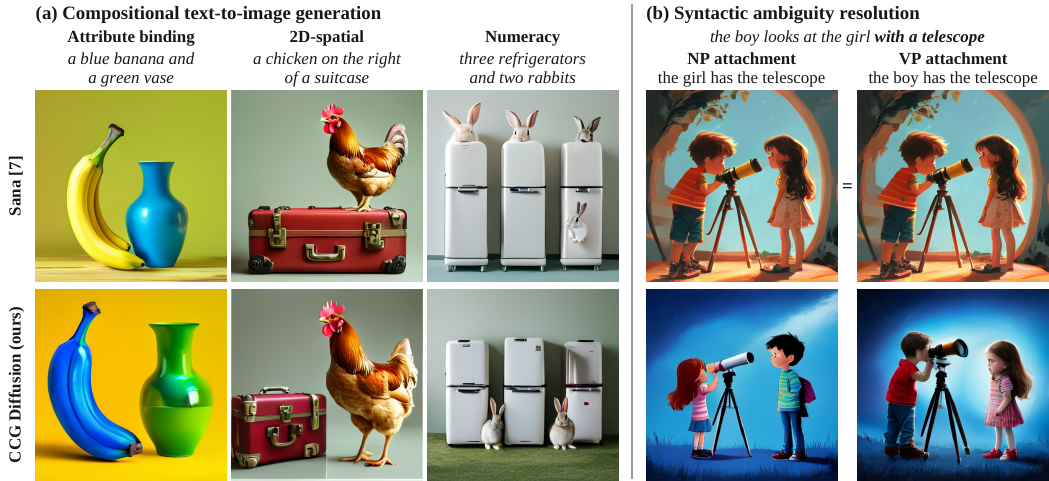}
\caption{\textbf{The two target problems addressed in this paper.} 
(a) Compositional text-to-image generation: generating an image that matches a prompt describing
multiple objects and their attributes, relations, and counts.
(b) Syntactic ambiguity resolution: generating an image that follows the
intended interpretation of a prompt that admits multiple syntactic structures,
e.g., whether a PP modifies a noun phrase (NP attachment) or a verb phrase
(VP attachment).
The proposed CCG Diffusion can address both of these problems.
}
\label{fig:teaser}
\end{figure*}

In text-to-image generation, the objects and attributes in a prompt must be
depicted not only individually but also faithfully to the syntactic structure
of the prompt, i.e., which word modifies which and where each phrase attaches.
We address two problems that require this syntactic faithfulness
(Fig.~\ref{fig:teaser}): compositional text-to-image generation and syntactic
ambiguity resolution. These problems persist even in recent text-to-image
diffusion models that generate high-quality
images~\cite{ramesh2022hierarchical,rombach2022high,podell2024sdxl,esser2024scaling,bfl2024flux,imagenteam2024imagen3,xie2025sana,wu2025qwenimage},
and they are part of a widely reported limitation: generated images are not always
faithful to what the prompt
describes~\cite{chefer2023attend,ghosh2023geneval,huang2025t2icompbenchpp,hayes2025finegrain,li2026easier}
or to the interpretation intended by the
user~\cite{mehrabi2023resolving,elsharif2025vlab,hahn2025proactive}.
Underlying these problems is the fact that the syntactic structure of the
prompt is not given to the model explicitly but is conveyed only implicitly
through the representations of the text encoder.

The first problem is compositional text-to-image generation: generating an image that
matches a prompt describing multiple objects and their attributes, relations,
and counts. Existing models exhibit failures such as binding attributes to the
wrong objects, fusing objects, and producing spatial relations or counts that
differ from the prompt. The top row of
Fig.~\ref{fig:teaser}(a) shows examples generated by Sana~\cite{xie2025sana}.
For \prompt{a blue banana and a green vase}, the banana remains yellow and the
vase is drawn blue instead of green; that is, \prompt{blue} is bound to
\prompt{vase}. For \prompt{a chicken on the right of a suitcase}, the chicken
stands on top of the suitcase rather than to its right. For \prompt{three
refrigerators and two rabbits}, four rabbits are drawn, some of them merged
with the refrigerators. All of these failures occur because the relations
expressed by the syntactic structure (which attribute word modifies which noun,
e.g., \prompt{blue} and \prompt{banana}; which relation phrase links which nouns,
e.g., \prompt{on the right of} with \prompt{chicken} and \prompt{suitcase}; and which
numeral applies to which noun, e.g., \prompt{two} and \prompt{rabbits}) are not
reflected in the relations among the objects in the image.

The second problem is syntactic ambiguity resolution: generating an image that follows
the intended interpretation of a prompt that admits multiple syntactic
structures. For example, in \prompt{the boy looks at the girl with a telescope}
in Fig.~\ref{fig:teaser}(b), the prepositional phrase (PP) \prompt{with a
telescope} can modify either the noun phrase \prompt{the girl} (NP attachment)
or the verb phrase \prompt{looks at the girl} (VP attachment). Under the
NP-attachment interpretation, the sentence describes a scene in which the girl
has the telescope; under the VP-attachment interpretation, the boy has it.
However, conventional models, which receive no explicit information about
syntactic structure, are given the same word sequence under both
interpretations; the user therefore cannot specify the intended
interpretation, and which interpretation is depicted is left to the
model. Indeed, as shown in the top row of
Fig.~\ref{fig:teaser}(b), Sana generates the same image under both
interpretations from the same initial noise, and this image depicts only the
VP-attachment scene, in which the boy has the telescope; a user who intends
the NP-attachment interpretation thus cannot obtain the desired image.
This problem remains important in recent systems in which a large language
model (LLM) or an LLM-based agent writes the prompt, for example from a
conversation with the
user~\cite{betker2023improving,hahn2025proactive,chen2025t2icopilot}. Such
long, detailed prompts tend to contain many PPs, the number of possible
syntactic structures grows combinatorially with the number of
PPs~\cite{church1982coping}, and LLMs themselves do not reliably recognize
ambiguity~\cite{liu2023afraid,karamolegkou2025trick}.

For compositional text-to-image generation, training-free attention guidance
updates the latent during sampling with the gradient of a loss on
cross-attention maps~\cite{chefer2023attend,agarwal2023astar,li2023divide,rassin2023linguistic,meral2024conform,zhang2024ebama}.
Predicated Diffusion~\cite{sueyoshi2024predicated}, on which our method is
based, is not a generative model but one such training-free method: it
derives its loss from propositions in predicate logic that represent the
intent of the prompt, thereby guiding the generation process of a pretrained
diffusion model toward satisfying these propositions. For syntactic ambiguity
resolution,
TIED~\cite{mehrabi2023resolving} concatenates disambiguating information
obtained from the user to the prompt. These methods have three limitations.
First, parsing-based methods use the parse mainly to extract attribute--object
pairs or noun phrases for attribute
binding~\cite{feng2023trainingfree,rassin2023linguistic,zhang2024ebama,trusca2024object,zhuang2024magnet,hu2024tome};
even the authors of Predicated Diffusion note that its
spaCy-based~\cite{honnibal2020spacy} extraction can miss the relation
expressed by \prompt{with} and suggest that more advanced parsing may be
needed~\cite{sueyoshi2024predicated}. Second, Predicated
Diffusion cannot express object positions or counts, and most other methods
that control them require inputs or modules beyond the prompt, such as boxes,
layouts generated by an LLM, counting networks, or trained layout-prediction
models~\cite{chen2024layoutguidance,xie2023boxdiff,lian2024llmgrounded,phung2024refocusing,kang2025counting,binyamin2025makeitcount}.
Third, for syntactically ambiguous prompts, the intended interpretation is
hard to reflect in generation. Parsing-based guidance follows the single parse
output by the parser and thus applies the same guidance under either
interpretation. TIED concatenates disambiguating information, such as
\prompt{the girl has the telescope}, to the prompt, but the ambiguous sentence
remains, so whether the image follows the intended interpretation is still left
to the text encoder; for example, the CLIP text encoder~\cite{radford2021learning}
used by Stable Diffusion represents word order and relations between words
poorly~\cite{yuksekgonul2023aro,kang2025clipideal}.

We therefore propose CCG Diffusion, a method that explicitly derives the
syntactic structure of a prompt with Combinatory Categorial Grammar
(CCG)~\cite{steedman2000syntactic} and reflects it in image generation.%
\footnote{The code is available at \url{https://github.com/ryosuke-yamaki/ccg-diffusion.git}.}
CCG is a grammar that assigns each word a category specifying what kinds of
adjacent words or phrases it combines with and what kind of phrase it forms.
In CCG, most of the structure of a sentence is determined by the categories of
its words~\cite{bangalore1999supertagging,clark2004importance}. Therefore,
which word modifies which and where each phrase attaches can be read directly
from the result of CCG parsing, called a
derivation. Moreover, because the
attachment site of a PP is determined by the category of the preposition alone, 
simply parsing with this category fixed yields a
derivation that follows the specified interpretation. The phrase-structure
(constituency) and dependency parsers used by the parsing-based methods discussed above
can also provide similar information, but less directly: extracting which
word modifies which requires additional rules tailored to their trees or
labels, and specifying an interpretation requires constraining the tree
structure itself during parsing. CCG, in contrast, handles both uniformly
through the category of each word.

CCG Diffusion conveys the syntactic information read from the derivation to
the image generation model through two pathways: rewriting the prompt given to
the text encoder, and attention guidance built on Predicated Diffusion. For
compositional text-to-image generation, it strengthens attribute binding,
encourages two objects with attributes to be drawn in different regions to
suppress attribute leakage and object fusion, and controls the spatial
relations and counts of objects. As a result, the bottom row of
Fig.~\ref{fig:teaser}(a) shows a blue banana and a green vase, a chicken to the
right of a suitcase, and three refrigerators with two rabbits, each image
faithfully reflecting what the prompt describes, exactly as its syntactic
structure specifies. For syntactic ambiguity resolution, we focus
on PP-attachment ambiguity, a representative case, and refer to its resolution
as PP ambiguity resolution. Using the derivation of the specified
interpretation, CCG Diffusion paraphrases the prompt into a sentence that
admits only that interpretation and guides the object of the PP to the region
of its holder under that interpretation; in the example above, the telescope
is guided to the girl under NP attachment and to the boy under VP attachment.
As a result, as shown in the bottom row of Fig.~\ref{fig:teaser}(b), CCG
Diffusion generates, from the same prompt, an image in which the girl has the
telescope under NP attachment and one in which the boy has it under VP
attachment, each faithfully reflecting the syntactic structure of the
specified interpretation.

CCG Diffusion requires no additional training of the parser or the generation
model, and uses no input other than the prompt and, for ambiguous prompts, the
specified interpretation. Because it acts only on the input text and the
generation process of a pretrained model, it can be applied to various
diffusion models with different architectures, as long as the attention maps
of text tokens are accessible. To the best of our knowledge, it is the first
method to use CCG to control text-to-image generation. We evaluate it on
benchmark datasets for both problems with two backbones: Stable Diffusion,
which consists of a U-Net and the CLIP text encoder and has been the standard
backbone for prior attention guidance methods, and Sana, a representative of
recent models that consist of a Diffusion Transformer
(DiT)~\cite{peebles2023scalable} and an LLM text encoder.

Our two main contributions are as follows:
\begin{enumerate}
  \item \textbf{CCG-based compositional text-to-image generation.} We
        extend Predicated Diffusion with syntactic information read from CCG
        derivations and improve control over attribute binding, spatial
        relations, and counts without using any input other than the prompt.
        In a quantitative evaluation on
        T2I-CompBench++~\cite{huang2025t2icompbenchpp}, CCG Diffusion achieves
        the highest scores among the compared methods in attribute binding and
        2D spatial relations on both backbones, Stable
        Diffusion~\cite{rombach2022high} and Sana~\cite{xie2025sana}, and also significantly
        outperforms Predicated Diffusion in numeracy.
  \item \textbf{PP ambiguity resolution by specifying the syntactic
        interpretation.} Using the CCG property that the attachment site of a
        PP is determined by the category of the preposition alone, we develop
        a method that obtains the derivation of the specified interpretation
        and reflects this interpretation in image generation through both
        prompt paraphrasing and attention guidance. In a quantitative
        evaluation on the PP-attachment prompts of the Text-to-image
        Ambiguity Benchmark (TAB)~\cite{mehrabi2023resolving}, CCG Diffusion
        significantly outperforms all compared methods, including one that
        concatenates the gold visual setup to the prompt, despite receiving
        only the specified interpretation. 
\end{enumerate}

\section{Related Work}
\label{sec:related}

\subsection{Compositional Text-to-Image Generation}
\label{sec:related-compositional}
Given prompts with multiple objects and attributes, text-to-image diffusion
models can omit some objects, bind attributes to the wrong objects, or fuse
multiple objects into one. Training-free attention guidance suppresses such
failures without retraining the model: it defines a loss on cross-attention
maps, which indicate the image region that each prompt token corresponds to,
and guides generation with the gradient of this loss. Most methods design
their losses for predefined types of failure, for example by strengthening the
attention of neglected objects~\cite{chefer2023attend,li2023divide} or by
keeping the attention maps of different objects
apart~\cite{agarwal2023astar,meral2024conform}; some edit the attention maps
or optimize the initial noise instead of the
latent~\cite{zhang2024attnreg,guo2024initno}. Predicated
Diffusion~\cite{sueyoshi2024predicated}, on which our work builds, instead
represents the intent of a prompt as propositions in predicate logic and
converts them into losses on attention maps.

Applying these methods requires the guidance targets, i.e., the words whose
attention is manipulated, and several methods determine them by parsing the
prompt. SynGen~\cite{rassin2023linguistic}, EBAMA~\cite{zhang2024ebama}, and
Trusca \textit{et al.}~\cite{trusca2024object} bind each attribute to the object
it modifies, using modifier--noun pairs found by dependency parsing, and
Predicated Diffusion extracts the targets of its propositions from the
part-of-speech tags and dependency labels of spaCy~\cite{honnibal2020spacy}.
In these methods, and in StructureDiffusion~\cite{feng2023trainingfree},
Magnet~\cite{zhuang2024magnet}, and ToMe~\cite{hu2024tome} described below, the
prompt is parsed with dependency or constituency parsers such as spaCy and
Stanza~\cite{qi2020stanza}, and the parse is used mainly to extract
attribute--object pairs or noun phrases for attribute binding. This
extraction relies on rules tailored to each parser's output, such as its
dependency labels or phrase types.

Methods that control the layout or number of objects by guiding generation
mostly require inputs other than the prompt or external modules: boxes or
target positions given by the
user~\cite{chen2024layoutguidance,xie2023boxdiff,epstein2023selfguidance},
layouts generated by an LLM~\cite{lian2024llmgrounded,phung2024refocusing}, an
external counting network~\cite{kang2025counting}, or a trained
layout-prediction model~\cite{binyamin2025makeitcount}.
In contrast, some methods control the spatial relations between objects
without boxes or layouts by guiding the attention centroids of the
objects~\cite{han2025storm,rastegar2025infsplign,izadi2025fine}.

Another line of work intervenes in the text conditioning given to the
denoiser. Composable Diffusion~\cite{liu2022compositional} composes the denoising scores
obtained under separate conditions;
StructureDiffusion~\cite{feng2023trainingfree} and
DisCLIP~\cite{trusca2024object} separately encode the constituents, such as
noun phrases, obtained by constituency parsing. Other methods merge or edit
token embeddings~\cite{hu2024tome,park2023energy,zhuang2024magnet},
or use an LLM, either through a trained connector~\cite{hu2024ella} or by
decomposing the prompt into rewritten subprompts, each assigned to an image
region planned by the LLM~\cite{yang2024rpg}.

\subsection{Syntactic Ambiguity Resolution}
\label{sec:related-ambiguity}
Syntactic ambiguity occurs when a sentence admits multiple syntactic
structures with different meanings; PP-attachment ambiguity is a typical
example. Since the LAVA corpus~\cite{berzak2015lava}, which pairs ambiguous
sentences with videos depicting each interpretation, most work has used visual
information to select the interpretation of a given sentence rather than to
generate images: for example, jointly resolving image segmentation and PP
attachment~\cite{christie2016resolving}, resolving PP attachment with the
objects in the image~\cite{delecraz2018flickr30kpp,delecraz2021multimodal},
and evaluating whether vision-language models can use images to resolve
ambiguity, which they still do not do
reliably~\cite{zhou2025focus,lee2026lavisa}.

On the generation side, studies have characterized ambiguity and
underspecification in text-to-image
generation~\cite{hutchinson2022underspecification}, compared the
interpretations depicted by diffusion models with those of
humans~\cite{elsharif2025vlab}, and examined through cross-attention how
models resolve ambiguous pronouns~\cite{park2024picturing}. These studies
mainly characterize ambiguity or analyze how models handle it.

Mehrabi \textit{et al.}~\cite{mehrabi2023resolving} built the Text-to-image Ambiguity
Benchmark (TAB) from the ambiguous sentences of LAVA and new examples,
covering syntactic ambiguities such as PP attachment as well as discourse
ambiguities. They also proposed TIED, which places a language model between
the user and the text-to-image model and resolves the ambiguity of a prompt
through interaction with the user. In QA-TIED, the user answers a clarifying
question generated by the language model; in VS-TIED, the user selects the
intended one from several visual setups, i.e., sentences that each describe a
possible interpretation. The answer or the selected visual setup is
concatenated to the original prompt and fed to the text-to-image model.
Hahn \textit{et al.}~\cite{hahn2025proactive} extended such clarification to
underspecified prompts with agents that ask questions and present their
estimate of the user's intent as an editable belief graph.

\section{Preliminaries}
\label{sec:preliminaries}

\subsection{Predicated Diffusion}
\label{sec:pd}
This subsection outlines Predicated Diffusion~\cite{sueyoshi2024predicated},
on which our method is based. Predicated Diffusion is an attention guidance
method that represents the intent of a prompt as predicate-logic
propositions. By treating cross-attention maps as fuzzy predicates, it
converts the degree to which the propositions are satisfied into a
differentiable loss.

Generation starts from pure noise $x_T$, and the state $x_t$ is denoised at
each step $t=T,\ldots,1$. In latent diffusion models such as Stable
Diffusion, this process operates on the latent obtained with a variational
autoencoder (VAE)~\cite{kingma2014auto} rather than on the image. For each word $w$ in the prompt
$c$, the cross-attention of the denoiser produces an attention map $A_w$. We
write $A_w[i]\in[0,1]$ for the value of $A_w$ at the $i$-th pixel.

Predicated Diffusion adds the gradient $-\nabla_{x_t}\mathcal{L}$ of a loss
$\mathcal{L}$ to the reverse-process update as guidance, steering the state
so that the generated image has a lower loss. That is, before each denoising
step, the state is updated as
\begin{equation}
  \widetilde{x}_t=x_t-\eta_t\nabla_{x_t}\mathcal{L},
  \label{eq:guidance}
\end{equation}
and one standard denoising step is applied to $\widetilde{x}_t$ to obtain
$x_{t-1}$.
Below, we omit the step index $t$.

The distinctive feature of Predicated Diffusion is that it derives this loss
$\mathcal{L}$ from propositions. It treats each word in the prompt as a
predicate and interprets the attention-map value $A_P[i]$ of word $P$ as the
degree to which the predicate holds at the $i$-th pixel (a fuzzy truth
value). For example, \prompt{banana} and \prompt{blue} in \prompt{a blue banana}
correspond to the predicates $\mathrm{Banana}(x)$ and $\mathrm{Blue}(x)$,
which mean ``there is a banana at this pixel'' and ``this pixel is blue,''
respectively. Predicated Diffusion then computes the truth value of a
proposition $R$ with fuzzy logic and uses its negative logarithm as the loss
$\mathcal{L}[R]$, thereby guiding the generated image to satisfy the
proposition.

Predicated Diffusion represents the intent of a prompt with the four types
of propositions in Table~\ref{tab:pd-losses}. Here, $N$, $M$, $O$, and $S$
are the predicates of words denoting a noun, a modifier, an object, and its
holder, respectively. Existence states that the object denoted by noun $N$ is
present somewhere in the image; correspondence, that modifier $M$ and the
noun $N$ it modifies appear at the same positions; leakage, that modifier $M$
does not appear at the positions of another noun $N$; and possession, that
holder $S$ is also present wherever object $O$ is present. These
propositions suppress missing objects, incorrect attribute binding,
attribute leakage, and missed possession relations, respectively. One
proposition is instantiated for each element of the sets in the Targets
column of Table~\ref{tab:pd-losses}. In the table, $\mathcal{E}$ is the set
of nouns that should be present, $\mathcal{C}$ is the set of pairs of a
modifier and the noun it modifies, $\mathcal{K}$ is the set of pairs of a
modifier and a noun it does not modify, and $\mathcal{P}$ is the set of
holder--object pairs. We collectively call them the guidance targets
$\mathcal{I}=(\mathcal{E},\mathcal{C},\mathcal{K},\mathcal{P})$. For example,
given the prompt \prompt{a blue banana and a green vase},
$\mathcal{E}=\{\prompt{banana},\prompt{vase}\}$,
$\mathcal{C}=\{(\prompt{blue},\prompt{banana}),\allowbreak(\prompt{green},\prompt{vase})\}$,
$\mathcal{K}=\{(\prompt{blue},\prompt{vase}),\allowbreak(\prompt{green},\prompt{banana})\}$,
and $\mathcal{P}$ is empty.

\begin{table}
\caption{\textbf{Defined proposition types in Predicated Diffusion.}}
\label{tab:pd-losses}
\setlength{\tabcolsep}{3pt}
\centering
\begin{tabular}{|l|l|l|}
\hline
Type & Targets & Proposition \\
\hline
Existence (exist) & $N\in\mathcal{E}$ & $\exists x.\,N(x)$ \\
Correspondence (corr) & $(M,N)\in\mathcal{C}$ & $\forall x.\,M(x)\leftrightarrow N(x)$ \\
Leakage (leak) & $(M,N)\in\mathcal{K}$ & $\forall x.\,M(x)\rightarrow\neg N(x)$ \\
Possession (poss) & $(S,O)\in\mathcal{P}$ & $\forall x.\,O(x)\rightarrow S(x)$ \\
\hline
\end{tabular}
\end{table}

For each proposition type
$\tau\in\{\mathrm{exist},\mathrm{corr},\mathrm{leak},\mathrm{poss}\}$, let
$\mathcal{R}_\tau$ be the set of propositions instantiated from the elements
of its targets in Table~\ref{tab:pd-losses}, and define their mean loss as
\begin{equation}
 \overline{\mathcal L}_\tau=\frac{1}{|\mathcal{R}_\tau|}\sum_{R\in\mathcal{R}_\tau}\mathcal{L}[R]
 \label{eq:pd-type}
\end{equation}
(set to zero when $\mathcal{R}_\tau$ is empty). The Predicated Diffusion loss
is their weighted sum
\begin{equation}
 \mathcal{L}_{\mathrm{PD}}
 =\gamma_{\mathrm{exist}}\overline{\mathcal L}_{\mathrm{exist}}
 +\gamma_{\mathrm{corr}}\overline{\mathcal L}_{\mathrm{corr}}
 +\gamma_{\mathrm{leak}}\overline{\mathcal L}_{\mathrm{leak}}
 +\gamma_{\mathrm{poss}}\overline{\mathcal L}_{\mathrm{poss}},
 \label{eq:pd-total}
\end{equation}
where $\gamma_{\mathrm{exist}}$, $\gamma_{\mathrm{corr}}$,
$\gamma_{\mathrm{leak}}$, and $\gamma_{\mathrm{poss}}$ are the weights of the
proposition types.

For prompts that follow a template, the guidance targets can be defined in
advance; otherwise, they can be extracted automatically by dependency
parsing. Predicated Diffusion extracts the guidance targets automatically
from the part-of-speech tags and dependency labels produced by spaCy.

\subsection{Combinatory Categorial Grammar}
\label{sec:holccg}
Combinatory Categorial Grammar (CCG)~\cite{steedman2000syntactic} is a
lexicalized grammar, which places most grammatical information in the
lexicon. CCG assigns each word a category that specifies the types and
directions of the arguments the word takes and the type that results from
combining with them. Categories consist of atomic categories, such as $N$
(noun), $NP$ (noun phrase), and $S$ (sentence), and functor categories built
from them. A category $X/Y$ takes an argument $Y$ on its right, and
$X\backslash Y$ takes an argument $Y$ on its left; both yield $X$. For example,
an attributive adjective has $N/N$, which takes a noun on its right to yield
a noun, and a transitive verb has $(S\backslash NP)/NP$, which takes the
object on its right and then the subject on its left to yield a sentence.
The categories of adjacent phrases are combined by a small set of
combinatory rules. The most basic rules are forward application ($>$), in
which $X/Y$ combines with $Y$ on its right to yield $X$, and backward
application ($<$), in which $X\backslash Y$ combines with $Y$ on its left to
yield $X$. Other rules include composition and coordination.

As a grammar for parsing, CCG has two characteristics that our method relies
on. First, predicate--argument dependencies between words can be read directly from the
parse~\cite{clark2007wide,hockenmaier2007ccgbank}, because the parsing
process records which word fills each argument slot of each functor category.
We use these dependencies to read which modifier modifies which noun, which
nouns a relation phrase links, and which noun a numeral counts, and thereby
determine the guidance targets (Section~\ref{sec:method-targets}). By
contrast, a phrase-structure tree does not mark such word-to-word relations,
which are usually recovered with head rules~\cite{collins2003head}. A
dependency tree marks them as labeled arcs, but deciding which arcs
correspond to these relations requires rules tailored to its label scheme, as
in Predicated Diffusion.
Second, because its rules are few and general, the categories of individual
words determine most of the sentence structure, and supertagging, which
assigns a category to each word, does most of the work of
parsing~\cite{bangalore1999supertagging,clark2004importance}. Hereafter, we
call the category assigned to each word its supertag. As a result, fixing the
supertag of a single word constrains the structure of the whole sentence. We
use this to obtain the derivation of a specified interpretation: for PP
attachment, fixing the supertag of the preposition alone determines the phrase
that the PP modifies (Section~\ref{sec:method-reading}). In phrase-structure
and dependency parsing, by contrast, the preposition has the same
part-of-speech tag under both interpretations, so specifying the
interpretation requires constraining the tree structure itself, such as the
constituent or head to which the PP attaches.

\begin{figure}[t]
\centering
\includegraphics{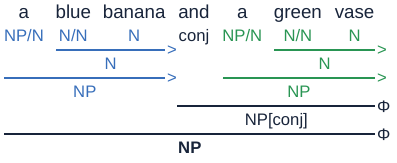}
\caption{\textbf{CCG derivation of an example sentence: \prompt{a blue banana and a green vase}.} The
category of each word is shown below it. Each line indicates that two adjacent
phrases are combined by the combinatory rule shown at its right end ($>$:
forward application; {\mathversion{normal}$\Phi$}: coordination), yielding the
category below the line. Blue and green mark the derivations of the two noun
phrases.}
\label{fig:ccg-coordination}
\end{figure}

The output of a CCG parser is called a derivation and is drawn as in
Fig.~\ref{fig:ccg-coordination}. 
The category of each word is placed below the word. Each horizontal line
indicates that the two adjacent phrases above it are combined, by the rule
shown at its right end, into a phrase with the category below the line. A
derivation is therefore read from top to bottom: from words to phrases and
from phrases to the sentence. For example, in
Fig.~\ref{fig:ccg-coordination}, the $N/N$ of \prompt{blue} combines with the
$N$ of \prompt{banana} on its right by forward application to yield $N$. This
$N$ then combines with the $NP/N$ of \prompt{a} to yield the $NP$ of the noun
phrase \prompt{a blue banana}. Likewise, \prompt{a green vase} becomes $NP$,
combines with the $\mathrm{conj}$ of \prompt{and} into $NP[\mathrm{conj}]$, and
then combines with the noun phrase on its left, so that the whole phrase
becomes a single $NP$ ($\Phi$). In this process, \prompt{banana} fills the
argument of the $N/N$ of \prompt{blue}, which naturally yields the word-to-word
dependency in which \prompt{blue} modifies \prompt{banana}. Similarly, in the
$(S\backslash NP)/NP$ of a transitive verb, the object fills the right $NP$
argument and the subject fills the left $NP$ argument, which yields the
verb--object and verb--subject dependencies.

\section{Method: CCG Diffusion}
\label{sec:method}
\begin{figure*}[p]
\centering
\includegraphics[height=8.5in]{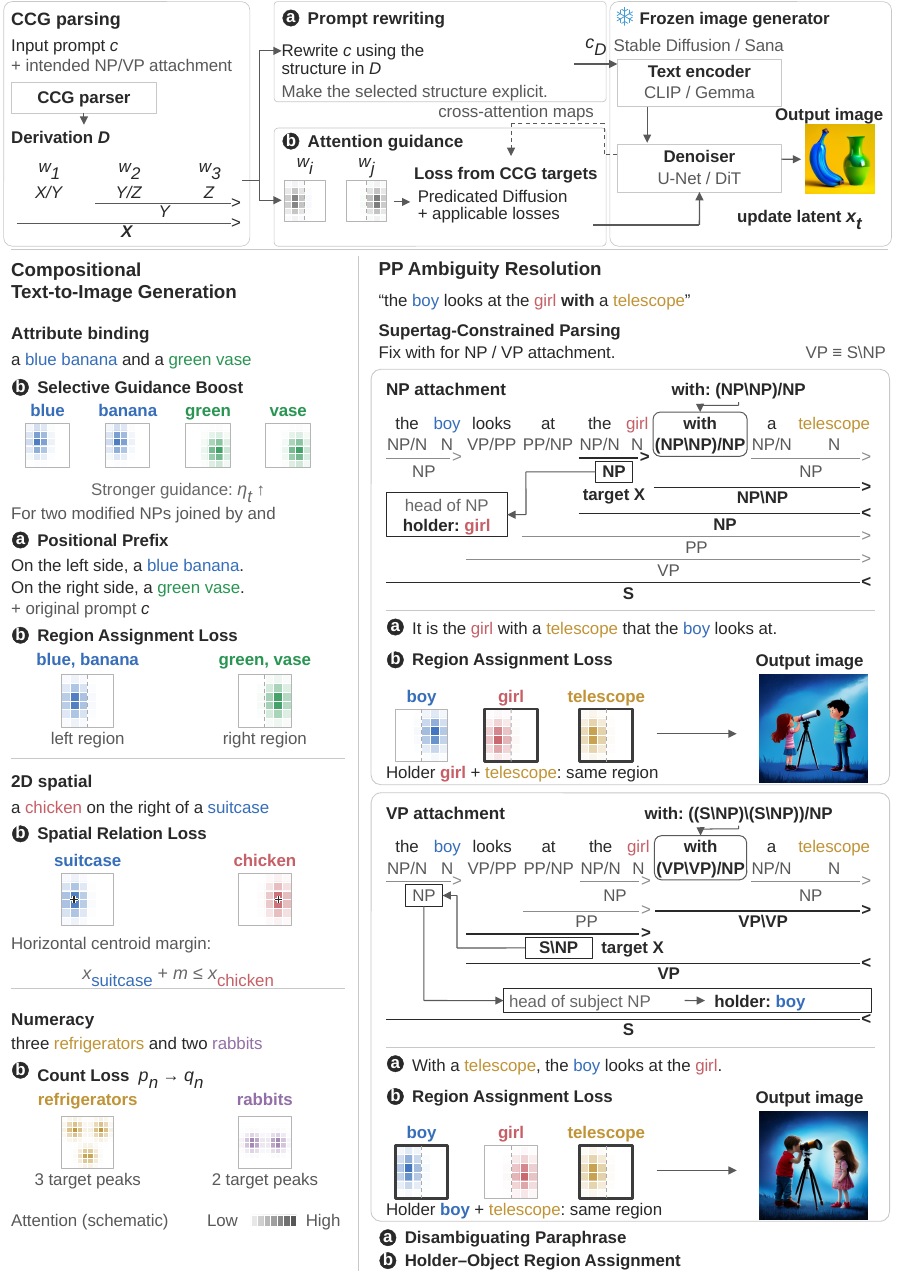}
\caption{\textbf{Overview of CCG Diffusion.} A CCG parser parses the prompt into a derivation $D$; for PP ambiguity
resolution, it parses the prompt under the specified NP or VP attachment. The
syntactic information read from $D$ is conveyed to the image generation model
through two pathways: (a) the input prompt to the text encoder and (b) attention
guidance.
}
\label{fig:pipeline}
\end{figure*}

In this section, we propose CCG Diffusion, which extends Predicated Diffusion
by integrating CCG parsing.
Fig.~\ref{fig:pipeline} gives an overview. CCG Diffusion parses the input
prompt with a CCG parser. Using the syntactic information in the resulting
derivation, it reduces errors in attribute binding, spatial relations, and
object counts in compositional text-to-image generation and disambiguates
PP attachment in syntactically ambiguous prompts according to the specified
interpretation. 
It achieves both in a unified manner and without additional training of 
the image generation model.

Let $D$ denote the derivation that the CCG parser outputs for the input
prompt $c$. CCG Diffusion uses the syntactic information in $D$ to control
generation through two pathways: prompt rewriting and attention guidance.
Below, we describe how it does so for the two problems, compositional
text-to-image generation and PP ambiguity resolution.

\subsection{Guidance Targets from CCG Derivation}
\label{sec:method-targets}
We first describe the guidance targets $\mathcal{I}_D$, which are shared by
compositional text-to-image generation and PP ambiguity resolution. In CCG
Diffusion, $\mathcal{I}_D=(\mathcal{E},\mathcal{C},\mathcal{K},\mathcal{P})$
is read from the categories and the predicate--argument dependencies in the
CCG derivation. The set $\mathcal{E}$ consists of the words with category $N$,
i.e., nouns. The set $\mathcal{C}$ consists of pairs of a word that modifies a
noun and the noun it modifies. For a word that modifies another modifier, we
follow the arguments and pair the word with the noun it ultimately modifies. For
example, in \prompt{a light blue banana}, \prompt{light} has category
$(N/N)/(N/N)$ and takes the $N/N$ of \prompt{blue} as its argument. Because
\prompt{blue} is not a noun, we further follow the argument of \prompt{blue} and
pair \prompt{light} with \prompt{banana}. This yields
$\mathcal{C}=\{(\prompt{light},\prompt{banana}),\allowbreak(\prompt{blue},\prompt{banana})\}$.
The set $\mathcal{K}$ pairs each modifier with the other nouns that it does
not modify. As in Predicated Diffusion, $\mathcal{P}$ consists of the
subject--object pairs of the verbs \prompt{have}, \prompt{wear}, \prompt{grasp},
and \prompt{hold}. The other grammatical structures used by the CCG Diffusion
components described below (coordinated noun phrases, relation phrases,
cardinal numbers, and PPs) are also read from the same CCG derivation.

\subsection{Compositional Text-to-Image Generation}
\label{sec:method-compositional}
This subsection addresses three of the common failure patterns in
compositional text-to-image generation: (1) attributes leak between objects
or objects fuse, (2) spatial relations are violated, and (3) object counts
are wrong. We mitigate these problems by extending Predicated Diffusion using
information from the CCG derivation.

\subsubsection{Selective Guidance Boost}
\label{sec:method-guidance-strength}
Attention guidance controls the generated image by updating the latent with
the loss gradient, as in \eqref{eq:guidance}; $\eta_t$ determines the size
of each update. A larger guidance strength enforces the
propositions of Predicated Diffusion more strongly and thus tends to improve
attribute binding. However, overly large updates can corrupt the
image~\cite{rassin2023linguistic}. We therefore replace the standard guidance
strength with a stronger one only for inputs in which the modifiers that the
CCG derivation binds to nouns include a word denoting color, shape, size,
material, or surface.

\subsubsection{Positional Prefix and Region Assignment Loss}
\label{sec:method-prefix-and-region}
When two noun phrases with modifiers are joined by \prompt{and}, image
generation models often fuse the objects or swap their attributes. We
mitigate this problem by separating the two objects into different regions of
the image. To this end, we combine a textual prefix that specifies where each
noun phrase should be placed with an attention guidance loss that constrains
where each object is generated.

\paragraph{Positional Prefix}
Text encoders such as CLIP, which are widely used in text-to-image models, are
generally causal: the state of a token depends only on the tokens before it. We
therefore place each noun phrase, preceded by a position cue, before the
original prompt:
\begin{equation}
 c_D=\text{\small\prompt{On the left side, }}\mathrm{NP}_1\text{\small\prompt{. On the right side, }}\mathrm{NP}_2\text{\small\prompt{.}}\,\|\,c,
 \label{eq:prefix}
\end{equation}
where $\|$ denotes concatenation, and NP$_1$ and NP$_2$ are the two noun
phrases in their order of appearance in the original prompt. We also keep the
original prompt to retain the context and the words outside the noun phrases.
For example, for the prompt \prompt{a blue banana and a green vase}, the text
encoder receives the following text:
\begin{center}\small
\begin{tabular}{@{}p{0.9\columnwidth}@{}}
\prompt{On the left side, a blue banana. On the right side, a green vase. a
blue banana and a green vase}
\end{tabular}
\end{center}
Because each noun phrase directly follows the phrase that indicates its
placement, the state of each noun is expected to reflect that placement. The
guidance losses are computed on the attention maps obtained from these
states. Methods that separately encode constituents such as noun phrases for
conditioning include
StructureDiffusion~\cite{feng2023trainingfree} and
DisCLIP~\cite{trusca2024object}. These methods insert the separately encoded
embeddings into the embedding of the original prompt or concatenate them with
it. In contrast, the Positional Prefix rewrites the input text itself. It
therefore applies in the same form to encoders such as CLIP and LLMs, without
manipulating the embeddings that the text encoder outputs.

\paragraph{Region Assignment Loss}
In addition to the Positional Prefix, we add an attention guidance loss that
assigns the two noun phrases to the left and right regions, respectively. To
represent the region where each object is placed along the horizontal axis of
the image, we use the smooth half-plane masks
\begin{equation}
 M_{\mathrm{L}}(x)=\operatorname{sigmoid}\bigl((x_{\mathrm{b}}-x)/\kappa\bigr),\quad
 M_{\mathrm{R}}(x)=1-M_{\mathrm{L}}(x).
 \label{eq:region-mask}
\end{equation}
Here, $x\in[0,1]$ is the horizontal position on the attention map (0 at the
left edge and 1 at the right edge), $x_{\mathrm{b}}$ is the position of the
left--right boundary, and $\kappa$ is a hyperparameter that sets the width
over which the mask value switches near the boundary.

Using these masks, we define the Region Assignment Loss as
\begin{equation}
 \mathcal L_{\mathrm{region}}
 =-\frac{1}{|\mathcal W_D|}\sum_{w\in\mathcal W_D}
 \log\frac{\sum_i M_{g(w)}[i]A_w[i]}{\sum_i A_w[i]},
 \label{eq:region-loss}
\end{equation}
where $\mathcal W_D$ is the set of nouns in the two noun phrases and their
modifiers, $A_w$ is the attention map of word $w$, and
$g(w)\in\{\mathrm{L},\mathrm{R}\}$ is the side assigned to the noun phrase
that contains $w$ (left for NP$_1$ and right for NP$_2$). The ratio inside the
logarithm is the fraction of the total attention of $w$ that falls within its
assigned mask. This loss belongs to the same family as the layout
guidance loss~\cite{chen2024layoutguidance}, which increases the fraction of
attention inside a specified region. It uses smooth half-plane masks instead
of boxes and, like the Predicated Diffusion losses, takes the negative
logarithm. However, whereas layout guidance and similar methods receive the
layout as boxes or as words that indicate positions, our method assigns left
and right sides to coordinated noun phrases whose positions are not
specified. We multiply $\mathcal L_{\mathrm{region}}$ by a weight
$\lambda_{\mathrm{region}}$ and add it to $\mathcal{L}_{\mathrm{PD}}$ in
\eqref{eq:pd-total}.

Applying the Positional Prefix and the Region Assignment Loss together
encourages the two objects to be drawn in separate left and right regions and,
as a result, mitigates attribute swapping and object fusion.

\subsubsection{Spatial Relation Loss}
\label{sec:method-2d}
The propositions defined in Predicated Diffusion do not depend on where
objects appear in the image, so they cannot express that one object is above,
below, to the left of, or to the right of another. The original Predicated
Diffusion paper~\cite{sueyoshi2024predicated} also lists binary predicates for
spatial relations, such as $\mathrm{Above}(x,y)$, as future work. For inputs
that specify a 2D spatial relation between two objects, we therefore introduce
a loss on the attention centroid of each object, following prior work that
guides object positions and spatial relations with attention
centroids~\cite{epstein2023selfguidance,han2025storm,rastegar2025infsplign,izadi2025fine}.
We detect a relation from the CCG derivation when one of a closed set of
relation phrases (\prompt{on the left of}, \prompt{on the right of}, \prompt{on the
top of}, \prompt{on the bottom of}, \prompt{next to}, \prompt{near}, and \prompt{on side
of}) links a subject noun $s$ and an object noun $o$.

Let $a_s$ and $a_o$ be the attention maps of the two nouns, and let
$(x_s,y_s)$ and $(x_o,y_o)$ be their attention-weighted centroids on the unit
square, with $y$ increasing downward. For a directional relation $\rho$, let
$d$ be the signed difference along the axis specified by $\rho$, defined to be
positive when $\rho$ holds ($d=x_o-x_s$ for \prompt{on the left of} and
$d=y_o-y_s$ for \prompt{on the top of}), and let $e$ be the difference along the
other axis. Then
\begin{equation}
 \begin{aligned}
 \ell^{\mathrm{dir}}_\rho&=[m-d]_+ + [|e|-|d|]_+,\\
 \ell^{\mathrm{prox}}_\rho&=[\min(|x_s-x_o|,|y_s-y_o|)-r]_+,\\
 \mathcal L_{\mathrm{spatial}}&=\frac{1}{|\mathcal R_D|}
 \sum_{\rho\in\mathcal R_D}\Bigl(\ell_\rho
 +\frac1I\sum_i a_s[i]a_o[i]\Bigr),
 \end{aligned}
 \label{eq:2d-loss}
\end{equation}
where $\mathcal R_D$ is the set of relations; $I$ is the number of pixels in
an attention map; and $\ell_\rho$ is $\ell^{\mathrm{dir}}_\rho$ for
\prompt{on the left of}, \prompt{on the right of}, \prompt{on the top of}, and
\prompt{on the bottom of}, and $\ell^{\mathrm{prox}}_\rho$ for \prompt{next to},
\prompt{near}, and \prompt{on side of}. The first hinge term requires a margin $m$
along the specified axis, the
second hinge term makes the difference along that axis exceed the difference
along the other axis, and the product term penalizes overlap so that the two
objects can be separated. Because the centroids lie on the unit square, $m$
and $r$ are fractions of the image size and are independent of the
resolution. 
As with $\mathcal L_{\mathrm{region}}$, we multiply
$\mathcal L_{\mathrm{spatial}}$ by a weight $\lambda_{\mathrm{spatial}}$ and
add it to $\mathcal{L}_{\mathrm{PD}}$ in \eqref{eq:pd-total}.

\begin{figure*}[t]
\centering
\includegraphics{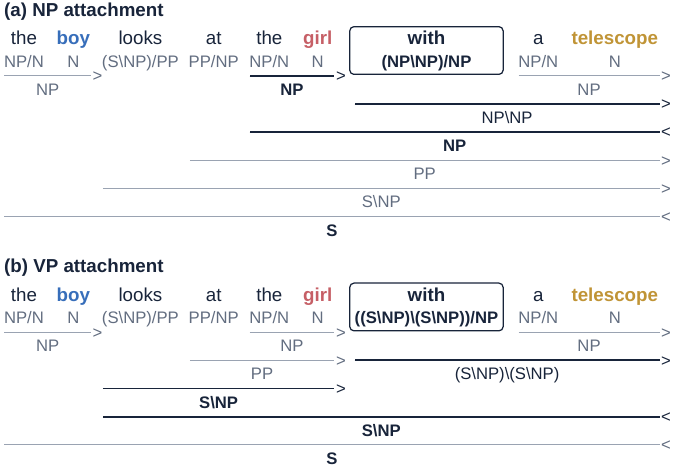}
\caption{\textbf{CCG derivations of an example sentence:}
\prompt{the boy looks at the girl with a
telescope} under the two interpretations. The derivations are obtained by
parsing the prompt with the CCG parser while fixing the supertag of \prompt{with}
(boxed) to (a) $(NP\backslash NP)/NP$ for NP attachment and (b)
$((S\backslash NP)\backslash(S\backslash NP))/NP$ for VP attachment.
}
\label{fig:ccg-pp}
\end{figure*}

\subsubsection{Count Loss}
\label{sec:method-count}
In Predicated Diffusion, the existence proposition $\exists x.\,N(x)$ is
satisfied as soon as a single object is present. It is therefore difficult to
generate an image with the specified number of objects, as in
\prompt{three refrigerators}. The original Predicated Diffusion
paper~\cite{sueyoshi2024predicated} also lists the inability to count objects
accurately as a limitation inherited from conventional diffusion models.
To address this, we introduce a loss that encourages the attention of a noun
to concentrate on as many mutually separated regions as the specified count,
instead of collapsing into one place. With each region corresponding to one
object, the loss aims to produce the specified number of objects.

First, we extract count constraints from the CCG derivation. We find each
noun modified by a cardinal number from \prompt{one} to \prompt{ten} and use the
pair $(n,k_n)$ of the noun and its count as a constraint. For example,
\prompt{three refrigerators and two rabbits} yields (\prompt{refrigerators}, 3)
and (\prompt{rabbits}, 2).

Next, for each noun, we build a target distribution $q_n$ that describes how
its attention should ultimately be distributed. The target $q_n$ has $k_n$
peaks at mutually separated locations in the image. We select the peaks from
highly attended locations in the noun's attention map such that they are
sufficiently far apart, aiming to obtain $k_n$ separate objects. We build
this target distribution only once, when the Count Loss is first applied,
from the attention before the latent is updated, and keep it fixed afterward.

After building the target, we pull the noun's attention distribution $p_n$,
smoothed and normalized to sum to 1, toward $q_n$ with the following loss:
\begin{equation}
 \begin{aligned}
 \mathcal L_{\mathrm{count}}
 ={}&\frac1N\sum_{n=1}^{N}g_n\,
 \frac12\sum_i\bigl(q_n[i]-p_n[i]\bigr)\\
 &\times\bigl(\log q_n[i]-\log p_n[i]\bigr),
 \end{aligned}
 \label{eq:count-loss}
\end{equation}
where $N$ is the number of nouns to be counted and $g_n$ is a weight that
prevents nouns with weak attention from being pushed too strongly. The sum
over $i$ is the symmetric KL divergence between $p_n$ and $q_n$. It penalizes
both missing attention at target peaks (too few objects) and attention
gathered where there is no peak (extra objects or collapse into one place).
As with $\mathcal L_{\mathrm{region}}$ and $\mathcal L_{\mathrm{spatial}}$, we
multiply $\mathcal L_{\mathrm{count}}$ by a weight $\lambda_{\mathrm{count}}$
and add it to $\mathcal{L}_{\mathrm{PD}}$ in \eqref{eq:pd-total}.

\subsection{PP Ambiguity Resolution}
\label{sec:method-ambiguity}
This subsection describes how we apply CCG Diffusion to syntactic ambiguity
resolution: given a syntactically ambiguous prompt, we generate an image that
corresponds to whichever of its syntactic interpretations is selected. Among
syntactic ambiguities, we focus on PP-attachment ambiguity, which is known as
a representative example.

CCG Diffusion takes as input a prompt with PP-attachment ambiguity together
with the interpretation intended for generation (NP attachment or VP
attachment). It then obtains the CCG derivation for this interpretation and,
based on the syntactic information in the derivation, extends Predicated
Diffusion to control generation so that the image corresponds to this
interpretation.

\subsubsection{Supertag-Constrained Parsing}
\label{sec:method-reading}
We obtain the derivation for the intended syntactic interpretation by running
the CCG parser with this interpretation specified, and we use the syntactic
information that the derivation contains.
In standard CCG parsing, the parser derives a single plausible syntactic
structure from the word-level CCG categories (supertags) that it predicts
itself. For a sentence with PP-attachment ambiguity, the supertag of its
preposition determines where the PP attaches. This follows from a property of
CCG as a lexicalized grammar: its supertags encode rich syntactic information.
Fig.~\ref{fig:ccg-pp} shows the derivations of the sentence \prompt{the boy
looks at the girl with a telescope} under its two interpretations, NP
attachment and VP attachment. As the figure shows, the
interpretation is NP attachment if the supertag of the preposition \prompt{with}
is $(NP\backslash NP)/NP$, and VP attachment if it is
$((S\backslash NP)\backslash(S\backslash NP))/NP$. Therefore, we parse the
prompt with the supertag of its preposition fixed to the one for the selected
interpretation, instead of letting the parser predict it. This allows us to
obtain the derivation for whichever interpretation is selected.

\subsubsection{Disambiguating Paraphrase and Holder--Object Region Assignment}
\label{sec:method-realization}
If a prompt with multiple syntactic interpretations is fed to the text encoder
as is, the diffusion model receives exactly the same conditioning for every
interpretation and thus no information about which interpretation is intended.
We therefore use the CCG derivation obtained for each interpretation to
paraphrase the prompt into a syntactically unambiguous sentence, and feed this
sentence to the text encoder instead of the original prompt. For NP
attachment, we rewrite the prompt as an object cleft sentence, \prompt{It is
[object NP] [PP] that [words before the object NP].} In this form, the PP lies
inside the focused noun phrase and can therefore modify only that noun. For VP
attachment, we front the PP, giving \prompt{[PP], [words before the PP].} In this
form, the PP is placed before the subject, away from the object noun, and
modifies the whole phrase. For example, given \prompt{the boy looks at the
girl with a telescope}, the text encoder receives \prompt{It is the girl with a
telescope that the boy looks at.} under the NP-attachment interpretation and
\prompt{With a telescope, the boy looks at the girl.} under the VP-attachment
interpretation. The idea of paraphrasing an ambiguous sentence into an
unambiguous one for each interpretation appears in the work of Duan
\textit{et al.}~\cite{duan2016generating}. We do not combine this paraphrase with the
Positional Prefix introduced in Section~\ref{sec:method-compositional}.

To also reflect the selected interpretation in the spatial layout of the
objects in the image, we use the Region Assignment Loss of
Section~\ref{sec:method-prefix-and-region}. We call the noun that is the object
of the PP the \emph{object} (\prompt{telescope} in the example), and the noun to
which the object belongs under the selected interpretation the \emph{holder}.
For NP attachment, the holder is the noun modified by the PP (\prompt{girl} in
the example); for VP attachment, it is the subject of the verb phrase modified
by the PP (\prompt{boy} in the example). That is, only the holder changes with
the interpretation. 
In the derivation, under either interpretation, the holder
and the object are obtained by the same rule from the arguments of the
preposition's supertag $(X\backslash X)/Y$. This supertag takes $Y$ on its
right as an argument and forms a phrase that modifies $X$ on its left. The
object is the noun of the noun phrase $Y$. The holder is the noun of the noun
phrase $X$ when $X=NP$; when $X=S\backslash NP$, it is the noun of the $NP$
that the verb phrase $X$ takes on its left, i.e., the subject noun. In the
Holder--Object Region Assignment, we assign the holder and the object of the
selected interpretation to the same region and the holder of the unselected
interpretation to the other region, and apply the Region Assignment Loss.
For example, under the NP-attachment interpretation of \prompt{the boy looks at
the girl with a telescope}, we guide \prompt{girl} and \prompt{telescope} to be
drawn in the same region and \prompt{boy} in the other region. Under the
VP-attachment interpretation, we guide \prompt{boy} and \prompt{telescope} to be
drawn in the same region and \prompt{girl} in the other region
(Fig.~\ref{fig:pipeline}). In this way, we use the Region Assignment Loss to
guide generation so that the syntactic interpretation is reflected in the
spatial layout of the objects in the image. In addition, we simultaneously
apply the Selective Guidance Boost (Section~\ref{sec:method-guidance-strength})
and the Spatial Relation Loss (Section~\ref{sec:method-2d}).

\section{Experiment 1: Compositional Text-to-Image Generation}
\label{sec:exp1}

\subsection{Setup}
\label{sec:exp1-setup}

\subsubsection{Dataset}
\label{sec:exp1-dataset}
To quantitatively evaluate our method on compositional text-to-image
generation, we conduct experiments on
T2I-CompBench++~\cite{huang2025t2icompbenchpp}. It is a standard benchmark
for this task that groups 8{,}000 prompts into four categories and eight
subcategories: attribute binding (color, shape, and texture), object
relationships (2D-spatial, 3D-spatial, and non-spatial), numeracy, and
complex compositions. Each subcategory has
1{,}000 prompts, which the benchmark splits into 700 training and 300 test
prompts. The prompts include both sentences based on fixed templates such as
\prompt{a \{adjective\} \{noun\} and a \{adjective\} \{noun\}} (e.g.,
\prompt{a blue banana and a green vase}) and natural sentences without
templates.

\nocite{gemmateam2024gemma2,yamaki2023holographic}
\begin{table*}[t]
\caption{\textbf{Performance on T2I-CompBench++.}
Higher is better in all categories. Bold: significantly higher than all
other rows of the same backbone. $^\dagger$: CCG Diffusion significantly
higher than Predicated Diffusion (paired permutation tests over the 300
prompts, Holm-corrected over the eight subcategories of each backbone,
$p<0.05$).}
\label{tab:exp1-main}
\setlength{\tabcolsep}{3pt}
\centering
\small
\begin{tabular*}{\textwidth}{@{\extracolsep{\fill}}|l|c|c|c|c|c|c|c|c|}
\hline
 & \multicolumn{3}{c|}{Attribute binding} & \multicolumn{2}{c|}{Spatial} & Numeracy & Non-spatial & Complex \\
Method & Color & Shape & Texture & 2D & 3D & (UniDet~\cite{zhou2022simple}) & (CLIPScore~\cite{hessel2021clipscore}) & (3-in-1) \\
\hline
\multicolumn{9}{|l|}{\textit{Stable Diffusion}} \\
\hline
Vanilla & 0.3794 & 0.3595 & 0.4169 & 0.1148 & 0.2957 & 0.4464 & 0.3108 & 0.2963 \\
Composable Diffusion~\cite{liu2022compositional} & 0.4198 & 0.3560 & 0.4301 & 0.1115 & 0.3056 & 0.4364 & 0.3085 & 0.3038 \\
StructureDiffusion~\cite{feng2023trainingfree} & 0.3587 & 0.3598 & 0.3987 & 0.1082 & 0.3039 & 0.4496 & 0.3093 & 0.3022 \\
Attend-and-Excite~\cite{chefer2023attend} & 0.5761 & 0.4802 & 0.6106 & 0.1043 & 0.3025 & 0.5191 & 0.3088 & 0.3279 \\
SynGen~\cite{rassin2023linguistic} & 0.7148 & 0.4889 & 0.6800 & 0.1195 & 0.3041 & 0.4485 & 0.3107 & 0.3384 \\
Predicated Diffusion~\cite{sueyoshi2024predicated} & 0.6631 & 0.4743 & 0.6221 & 0.1279 & 0.3341 & 0.5214 & 0.3105 & 0.3420 \\
CCG Diffusion (ours) & \textbf{0.7419}$^\dagger$ & \textbf{0.5582}$^\dagger$ & \textbf{0.7037}$^\dagger$ & \textbf{0.3051}$^\dagger$ & 0.3335 & \textbf{0.5365}$^\dagger$ & 0.3105 & 0.3467 \\
\hline
\multicolumn{9}{|l|}{\textit{Sana}} \\
\hline
Vanilla & 0.6796 & 0.5250 & 0.6413 & 0.3135 & 0.4051 & 0.6082 & 0.3133 & 0.3717 \\
Composable Diffusion & 0.4344 & 0.3588 & 0.4480 & 0.3117 & 0.4054 & 0.5291 & 0.3115 & 0.3658 \\
StructureDiffusion & 0.4757 & 0.3942 & 0.5247 & 0.2873 & 0.4004 & 0.5732 & 0.3127 & 0.3655 \\
Attend-and-Excite & 0.6904 & 0.5623 & 0.6689 & 0.3311 & 0.4149 & \textbf{0.6434} & 0.3128 & 0.3752 \\
SynGen & 0.8015 & 0.5689 & 0.7086 & 0.3120 & 0.4054 & 0.6069 & 0.3134 & 0.3800 \\
Predicated Diffusion & 0.7212 & 0.5490 & 0.6693 & 0.3304 & 0.4168 & 0.6151 & 0.3133 & 0.3776 \\
CCG Diffusion (ours) & 0.8072$^\dagger$ & \textbf{0.6196}$^\dagger$ & 0.7217$^\dagger$ & \textbf{0.4525}$^\dagger$ & 0.4168 & 0.6303$^\dagger$ & 0.3133 & 0.3777 \\
\hline
\end{tabular*}
\end{table*}

This benchmark is suitable for evaluating our method because its
subcategories directly correspond to the three failure patterns addressed in
Section~\ref{sec:method-compositional}. Attribute binding corresponds to
attribute leakage and object fusion, 2D-spatial to violated spatial
relations, and numeracy to incorrect object counts. We can thus measure the
effect of the extension for each pattern
(Sections~\ref{sec:method-guidance-strength}--\ref{sec:method-count})
separately, using the automatic evaluator of the corresponding subcategory. In addition,
because the benchmark contains natural sentences as well as template-based
prompts, we can evaluate our method on inputs that are not limited to fixed
sentence patterns.

For each subcategory, we use the official 300-prompt test split, generate
images with ten seeds per prompt, and report results obtained with the
official automatic evaluation procedure.

\subsubsection{Backbone Image Generation Models and CCG Parser}
\label{sec:exp-models}
We use two different text-to-image models as the backbones of our method and
the compared methods.

The first is Stable Diffusion~v1.4~\cite{rombach2022high} (Stable Diffusion
below), a representative latent diffusion model that consists of the CLIP text
encoder~\cite{radford2021learning} and a U-Net. We use the
\path{CompVis/stable-diffusion-v1-4}%
\footnote{\url{https://huggingface.co/CompVis/stable-diffusion-v1-4}}
checkpoint. Stable Diffusion has been the standard backbone for existing
attention guidance methods, and we also use it for a fair comparison.

The second is Sana~0.6B~\cite{xie2025sana} (Sana below), a fast, high-quality
model with a relatively recent architecture based on the Diffusion Transformer
(DiT)~\cite{peebles2023scalable} instead of a U-Net. We use the
\path{Efficient-Large-Model/}\allowbreak\path{Sana_600M_512px_diffusers}%
\footnote{\url{https://huggingface.co/Efficient-Large-Model/Sana_600M_512px_diffusers}}
checkpoint. Sana combines a Gemma~2~\cite{gemmateam2024gemma2} text encoder with
a DiT that uses linear attention. It has neither a U-Net nor the CLIP text
encoder, and using it as a backbone lets us verify that our method is not
tailored to a particular backbone model.

For the CCG parser of our method, we use
Holographic CCG (Hol-CCG)~\cite{yamaki2023holographic} with the
\path{ryosuke-yamaki/}\allowbreak\path{hol-ccg-}\allowbreak\path{roberta-large}%
\footnote{\url{https://huggingface.co/ryosuke-yamaki/hol-ccg-roberta-large}}
checkpoint. Hol-CCG is one of the state-of-the-art neural CCG parsers that
use a pre-trained language model as their text encoder.

\subsubsection{Compared Methods}
We compare our method with five training-free methods that, like ours, use no
input other than the prompt: the four methods compared in the Predicated
Diffusion paper~\cite{sueyoshi2024predicated} and Predicated Diffusion
itself. Besides keeping continuity with the evaluation of our base method,
this set matches the design of CCG Diffusion, which conveys syntactic
information through two pathways, the text conditioning and attention
guidance, and uses a parse to decide which words each intervention acts on.
The compared methods cover both pathways, each with and without syntactic
parsing, so that our method is compared with a prompt-only alternative for
each of these design choices:
\begin{itemize}
  \item Composable Diffusion~\cite{liu2022compositional}: intervenes in the
  conditioning without parsing, by composing the predictions of the denoiser
  obtained under separate conditions. We use the parts of the prompt separated
  by \prompt{and} as the conditions.
  \item StructureDiffusion~\cite{feng2023trainingfree}: intervenes in the text
  conditioning on the basis of constituency parsing, by separately encoding
  the noun phrases of the prompt.
  \item Attend-and-Excite~\cite{chefer2023attend}: applies attention guidance
  without syntactic parsing, by strengthening the attention of the most
  neglected of the target nouns. We use all nouns in the prompt as the
  targets.
  \item SynGen~\cite{rassin2023linguistic}: applies attention guidance on the
  basis of dependency parsing, by aligning the attention maps of
  modifier--noun pairs; it is specialized in attribute binding.
  \item Predicated Diffusion~\cite{sueyoshi2024predicated}: applies attention
  guidance with propositions extracted from dependency parsing. Since our
  method extends it, the comparison with it shows the effect of our
  CCG-based extensions.
\end{itemize}
We exclude methods that require inputs or modules other than the prompt, such
as boxes, layouts generated by an LLM, or counting networks
(Section~\ref{sec:related-compositional}), because they use information that
our method does not receive. Other prompt-only methods mostly refine the
attention-guidance losses of the above methods, for example for neglected or
overlapping objects~\cite{li2023divide,agarwal2023astar,meral2024conform} or
for modifier--noun binding~\cite{li2023divide,meral2024conform,zhang2024ebama};
we therefore use one
representative of each approach.

None of these methods provides an implementation for Sana; we therefore
adapted each of them to Sana, using the same inputs and generation settings as
for our method.

\begin{table*}[t]
\caption{\textbf{Score drop from removing each component on attribute binding in T2I-CompBench++.}
Score with the listed component disabled minus the score with all three
components enabled (two-sided paired permutation test over the 300 prompts,
Holm-corrected over the nine comparisons per backbone). * $p<0.05$,
** $p<0.01$, *** $p<0.001$, n.s.\ not significant.}
\label{tab:ablation-t2i-effect}
\setlength{\tabcolsep}{4pt}
\centering\small
\begin{tabular*}{\textwidth}{@{\extracolsep{\fill}}l|ccc|ccc}
\hline
 & \multicolumn{3}{c|}{Stable Diffusion} & \multicolumn{3}{c}{Sana}\\
Removed component & Color & Shape & Texture & Color & Shape & Texture\\
\hline
Selective Guidance Boost & $-$0.032\textsuperscript{***} & $-$0.041\textsuperscript{***} & $-$0.048\textsuperscript{***} & $-$0.012\textsuperscript{**} & $-$0.010\textsuperscript{*} & $-$0.031\textsuperscript{***}\\
Positional Prefix & $-$0.021\textsuperscript{**} & $-$0.035\textsuperscript{***} & $-$0.016\textsuperscript{***} & $-$0.065\textsuperscript{***} & $-$0.044\textsuperscript{***} & $-$0.022\textsuperscript{***}\\
Region Assignment Loss & $-$0.030\textsuperscript{***} & $-$0.014\textsuperscript{**} & $-$0.011\textsuperscript{**} & $-$0.006\textsuperscript{n.s.} & $-$0.002\textsuperscript{n.s.} & $-$0.003\textsuperscript{n.s.}\\
\hline
\end{tabular*}
\end{table*}

\subsection{Results}
\label{sec:exp1-results}
Table~\ref{tab:exp1-main} reports the results on T2I-CompBench++.
In the attribute-binding categories, our method achieves the highest scores
among all methods on both backbones. It significantly outperforms all other
methods in all three categories on Stable Diffusion and in shape on Sana. For
color and texture on Sana, the margin over SynGen is small and not
significant. Our method also significantly improves over Predicated Diffusion,
on which it is based, in all three categories on both backbones.
However, the contribution of each component of CCG Diffusion to performance
differs between the backbones, as we discuss in
Section~\ref{sec:ablation-compositional}.

In the 2D-spatial category, our method outperforms all other methods
significantly and by a large margin on both backbones. This shows that the
Spatial Relation Loss introduced in our method is effective.

In the numeracy category, our method yields a small but significant
improvement over Predicated Diffusion on both backbones. On Stable Diffusion
in particular, it significantly outperforms all other methods.

These results show that our method significantly outperforms its base method,
Predicated Diffusion, on both backbones in every compositional text-to-image
generation challenge it targets (attribute binding, 2D-spatial, and
numeracy). Its effectiveness is thus confirmed on two diffusion models with
different architectures. Moreover, in the 3D-spatial, non-spatial, and
complex categories, which our extensions do not primarily target, there is no
significant difference between our method and Predicated Diffusion. That is,
at least in these categories, we observe no degradation in generation
performance.

\begin{figure*}[p]
\centering
\includegraphics[height=8.5in]{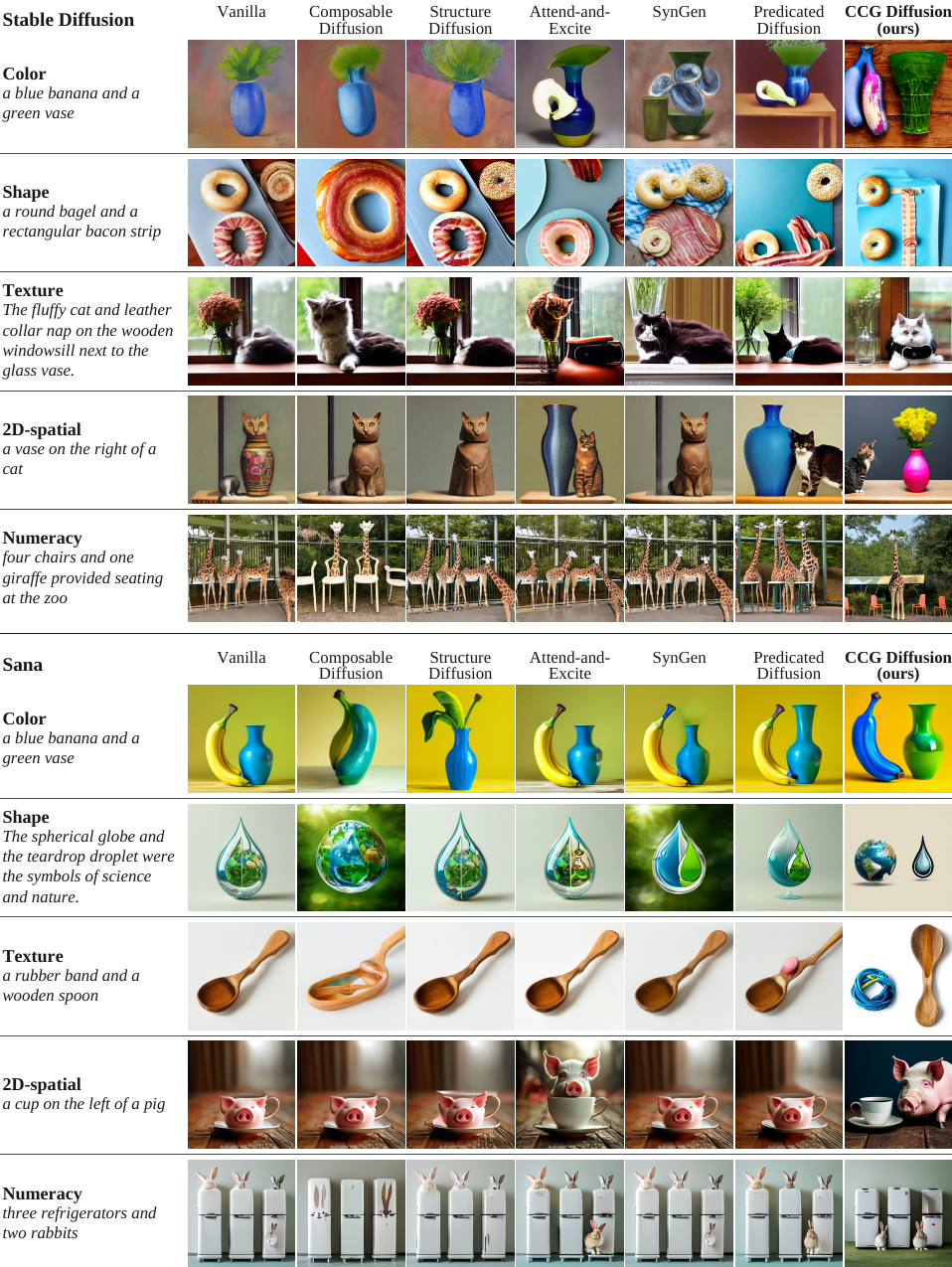}
\caption{\textbf{Qualitative comparison on Stable Diffusion (top) and Sana (bottom)
for the Color, Shape, Texture, 2D-spatial, and Numeracy categories.}}
\label{fig:t2i-qual}
\end{figure*}

Fig.~\ref{fig:t2i-qual} shows example images generated by the compared
methods and our method in the attribute-binding, 2D-spatial, and numeracy
categories. These are samples in which the compared methods fail to depict
the attribute binding, the spatial relations between objects, or the object
counts, whereas our method succeeds.
Appendix~\ref{app:qualitative} provides more generated samples, as well as
samples for the 3D-spatial, non-spatial, and complex categories.

\begin{table}[t]
\caption{\textbf{Score drop from removing the Spatial Relation Loss and the Count Loss in T2I-CompBench++.}
T2I-CompBench++ score with each objective removed minus the score of CCG Diffusion, evaluated
on the category that the objective targets (two-sided paired permutation test
over the 300 prompts, Holm-corrected over the two comparisons per backbone).
* $p<0.05$, ** $p<0.01$, *** $p<0.001$, n.s.\ not significant.}
\label{tab:ablation-t2i-single}
\setlength{\tabcolsep}{4pt}
\centering\small
\begin{tabular}{l|c|c}
\hline
Removed objective (category) & Stable Diffusion & Sana\\
\hline
Spatial Relation Loss (2D) & $-$0.177\textsuperscript{***} & $-$0.122\textsuperscript{***}\\
Count Loss (numeracy) & $-$0.013\textsuperscript{**} & $-$0.005\textsuperscript{n.s.}\\
\hline
\end{tabular}
\end{table}

\subsection{Ablation Study}
\label{sec:ablation-compositional}
We conduct an ablation study to isolate the contribution of each main
component of our method to the overall performance.
\subsubsection{Attribute Binding}
We first examine the main components for improving attribute binding: the
Selective Guidance Boost (Section~\ref{sec:method-guidance-strength}), the
Positional Prefix, and the Region Assignment Loss
(Section~\ref{sec:method-prefix-and-region}). We prepare variants in which each
of these components is independently enabled or disabled, and evaluate them on
the attribute-binding categories of T2I-CompBench++.
Table~\ref{tab:ablation-t2i-effect} shows how the score changes when each
component is disabled in the configuration with all three components, and
Appendix~\ref{app:ablation-full} shows the changes when two or more components
are disabled.

Disabling any one of the three components lowers the score in every category
on both backbones.
Disabling either the Selective Guidance Boost or the Positional Prefix
significantly lowers the score in every category on both backbones.
However, the size of each component's contribution differs between the
backbones.
Specifically, the Selective Guidance Boost and the Region Assignment Loss,
which act on attention guidance, tend to contribute more on Stable Diffusion,
whereas the Positional Prefix, which acts on the input to the text encoder,
tends to contribute more on Sana.
One possible reason is the difference in representational capacity between
the text encoders.
The CLIP text encoder of Stable Diffusion learns whole-sentence embeddings with
a contrastive objective, and such embeddings have been reported to be unable to
represent attribute--object binding and spatial relations
correctly~\cite{kang2025clipideal}.
The per-token states that Stable Diffusion uses for conditioning are learned
only indirectly through this objective, so they likely represent these
relations poorly as well.
This likely explains two observations on Stable Diffusion: the Positional
Prefix, which rewrites the input text, contributes significantly but, for
color, only about one third as much as on Sana; and guidance that acts
directly on attention contributes more.
In contrast, Gemma, the text encoder of Sana, is an LLM trained on large-scale
text corpora and can readily represent the placement instructions given by the
Positional Prefix and the attribute binding within each noun phrase.
Disabling the Region Assignment Loss on Sana causes no significant drop,
possibly because the Positional Prefix alone readily separates the objects,
leaving little room for guidance, particularly the Region Assignment Loss, to
complement it.
In short, we believe that the Selective Guidance Boost and the Region
Assignment Loss tend to contribute more, particularly on backbones whose text
encoders have weak representational capacity, and that the Positional Prefix
tends to contribute more on backbones whose text encoders have strong
representational capacity.

\subsubsection{2D-Spatial and Numeracy}
Next, Table~\ref{tab:ablation-t2i-single} shows how removing the Spatial
Relation Loss (Section~\ref{sec:method-2d}) or the Count Loss
(Section~\ref{sec:method-count}) from our method changes the score in the
category that each objective targets: 2D-spatial and numeracy, respectively.
Removing the Spatial Relation Loss substantially lowers the 2D-spatial score on
both backbones, and the drop is larger on Stable Diffusion.
As with attribute binding, the larger drop on Stable Diffusion may be partly
due to the difference in representational capacity between the text encoders
of the two backbones.
However, the drop on Sana is still large. Even with a strong text encoder,
reflecting the direction of a relation in the image remains difficult, so an
objective that explicitly constrains where objects are drawn is likely
effective.

In contrast, removing the Count Loss causes a small but significant drop in the
numeracy score on Stable Diffusion, whereas the drop on Sana is not
significant. Splitting the attention of a noun into the specified number of
regions does not guarantee that the image shows that many objects, and
changing the number of objects is a larger structural change than moving them.
Attention manipulation alone is therefore likely insufficient to control the
number of objects.
Both objectives contribute more on Stable Diffusion, which is consistent with
the trend observed for attribute binding: intervention through attention
guidance contributes more on Stable Diffusion.

\section{Experiment 2: PP Ambiguity Resolution}
\label{sec:exp2}

\subsection{Setup}
\label{sec:exp2-setup}

\subsubsection{Dataset}
To evaluate PP ambiguity resolution, we use the PP-attachment category of the
Text-to-image Ambiguity Benchmark (TAB)~\cite{mehrabi2023resolving}.
For each ambiguous prompt, TAB provides visual setups, each describing one
possible interpretation as a declarative sentence, and intention questions,
each asking about one visual setup as a yes/no question. For example, the
prompt \prompt{the boy looks at the girl with a telescope} comes with the visual
setups \prompt{the girl has the telescope} and \prompt{the boy has the telescope},
and the intention questions \prompt{does the girl have telescope} and \prompt{does
the boy have telescope}. Each intention question is designed so that its answer
is Yes if the image follows the interpretation of the corresponding visual
setup. The TAB paper proposes a framework that automatically evaluates whether
a generated image depicts the specified interpretation by feeding the image and
an intention question to a visual question answering (VQA) model, which
answers Yes or No. We adopt the
same evaluation protocol. To make the evaluation more reliable, we use two
different, independent LLMs as VQA models and cross-check their results.
Specifically, we use GPT-6 Astra~\cite{openai2026gpt6astra} (reasoning effort:
high) and Qwen3.8-Flash-Next~\cite{qwen2026design} (reasoning effort: xhigh) as
a closed and an open LLM, respectively, both with advanced reasoning
capabilities on image inputs. For Qwen3.8-Flash-Next, we use the
NVFP4-quantized checkpoint released by NVIDIA.%
\footnote{\url{https://huggingface.co/nvidia/Qwen3.8-Flash-Next-NVFP4}}

\begin{figure*}[!t]
\centering
\includegraphics[width=\textwidth]{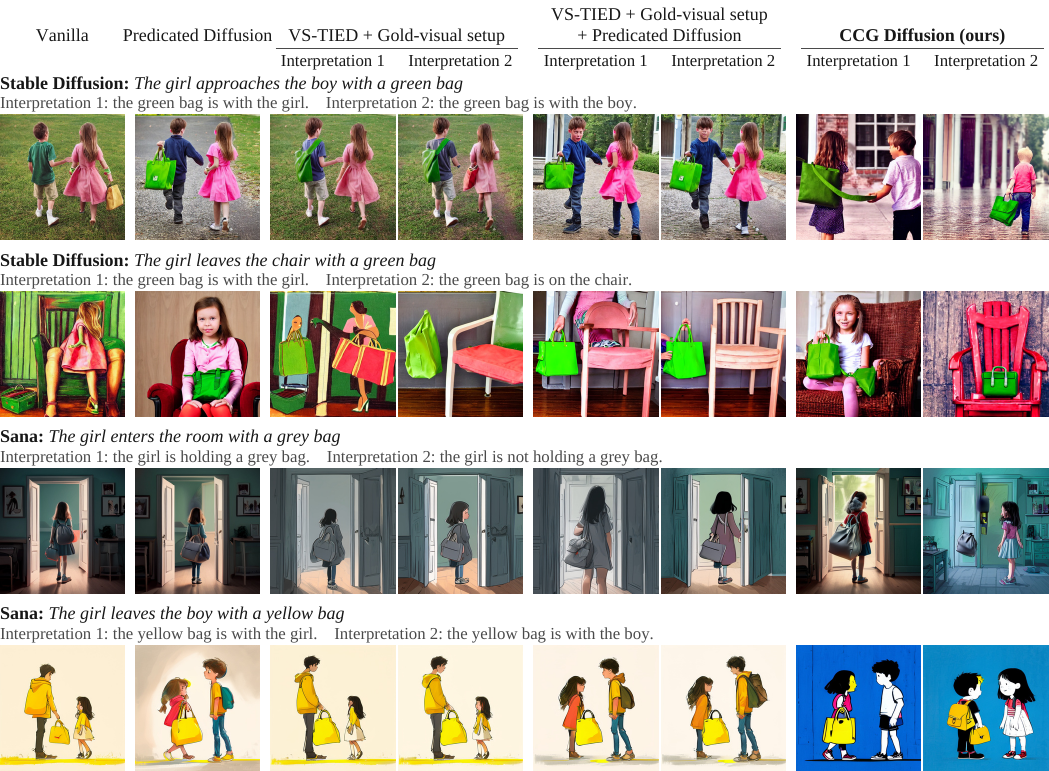}
\caption{\textbf{Examples of generated images for the TAB PP-attachment category on Stable Diffusion and Sana.}}
\label{fig:tab-qual-main}
\end{figure*}

For the 74 ambiguous prompts in the PP-attachment category, we generate images
with five seeds for each interpretation of each prompt. As the quantitative
metric, we report the Yes rate, i.e., the proportion of generated images for
which each VQA model answers Yes.

\subsubsection{Backbone Image Generation Models and CCG Parser}
As in Section~\ref{sec:exp-models}, we use Stable Diffusion and Sana as the
backbone text-to-image models and Hol-CCG as the CCG parser, and report the
performance on both backbones.

\subsubsection{Compared Methods}
We compare our method with the following methods:
\begin{itemize}
  \item Vanilla: We feed each ambiguous PP-attachment prompt as is to the
  backbone text-to-image model. Because the prompt is not disambiguated, the
  model receives no information about the intended interpretation, and its
  output does not differ between interpretations at all. Therefore, we generate
  only one image per seed and evaluate it with the intention questions of both
  interpretations.
  \item Predicated Diffusion: We feed each ambiguous prompt as is to Predicated
  Diffusion, on which our method is based. Predicated Diffusion takes no
  interpretation as input and extracts propositions from the single dependency
  parse that spaCy outputs for the prompt, so its guidance targets are the
  same regardless of the interpretation. Therefore, as with vanilla, its output
  does not differ between interpretations, and we evaluate the single image per
  seed with the intention questions of both interpretations.
  \item VS-TIED + Gold-visual setup:
  TIED~\cite{mehrabi2023resolving} resolves ambiguity at the prompt level: it
  elicits disambiguating information from the user for an ambiguous prompt and
  concatenates it to the original prompt. In its VS-TIED variant, the user
  selects the intended visual setup from candidate visual setups generated by a
  language model. In this experiment, we skip the generation by the language
  model and the selection by the user. We directly concatenate the ground-truth
  visual setup (gold visual setup) of each interpretation provided in TAB to the
  original prompt and feed the resulting prompt to the backbone. As the
  backbone, we use the same text-to-image model as in vanilla. We generate
  images from a different prompt for each interpretation and evaluate each image
  with the intention question of that interpretation.
  \item VS-TIED + Gold-visual setup + Predicated Diffusion: As in VS-TIED +
  Gold-visual setup, we concatenate the gold visual setup to the prompt, and
  then apply the
  Predicated Diffusion method above. This method combines text-level
  disambiguation of the prompt with the attention guidance of the existing
  Predicated Diffusion.
\end{itemize}

\subsection{Results}
\label{sec:exp2-results}
Table~\ref{tab:exp2-main} shows the results on TAB. On both backbones and in
the automatic evaluations by both LLMs, our method significantly outperforms
all other methods, including those that use the gold visual setup. This shows
that the combination of the Disambiguating Paraphrase and the Region
Assignment Loss, both obtained from the derivation of the specified
interpretation, is effective for rendering the interpretations of a
PP-attachment ambiguity as distinct images.
Moreover, the significant improvements on both backbones, which have
different architectures, suggest that the effectiveness of our method is not
limited to a particular model architecture.

\begin{table}[!t]
\caption{\textbf{TAB PP comparison on Stable Diffusion and Sana.}
Bold: significantly higher than all other rows of the same backbone under the
same VQA model. $^\dagger$: CCG Diffusion significantly higher than
Predicated Diffusion (paired permutation test over the 74 prompts,
Holm-corrected over the two VQA models per backbone, $p<0.05$).
Cohen's $\kappa$: agreement between the judgments of the two VQA models over
the 740 trials of each row.}
\label{tab:exp2-main}
\setlength{\tabcolsep}{2.5pt}
\centering\small
\begin{tabular}{|l|r|r|r|}
\hline
 & \multicolumn{2}{c|}{Yes rate (\%)} & Cohen's\\
\cline{2-3}
Method & GPT-6 & Qwen3.8 & \multicolumn{1}{c|}{$\kappa$}\\
\hline
\multicolumn{4}{|l|}{\textit{Stable Diffusion}}\\
\hline
Vanilla & 63.0 & 55.3 & 0.62\\
Predicated Diffusion & 66.9 & 62.2 & 0.61\\
VS-TIED + Gold-visual setup & 70.0 & 60.0 & 0.58\\
\begin{tabular}{@{}l@{}}VS-TIED + Gold-visual setup\\\quad+ Predicated Diffusion\end{tabular} & 67.2 & 63.5 & 0.72\\
CCG Diffusion (ours) & \textbf{75.3}$^\dagger$ & \textbf{70.9}$^\dagger$ & 0.72\\
\hline
\multicolumn{4}{|l|}{\textit{Sana}}\\
\hline
Vanilla & 64.1 & 60.7 & 0.79\\
Predicated Diffusion & 66.4 & 60.1 & 0.79\\
VS-TIED + Gold-visual setup & 70.9 & 67.7 & 0.80\\
\begin{tabular}{@{}l@{}}VS-TIED + Gold-visual setup\\\quad+ Predicated Diffusion\end{tabular} & 73.5 & 68.2 & 0.79\\
CCG Diffusion (ours) & \textbf{82.7}$^\dagger$ & \textbf{78.6}$^\dagger$ & 0.78\\
\hline
\end{tabular}\par
\vspace{-3pt}
\end{table}

We also measure the agreement between the judgments of the two VQA models
with Cohen's $\kappa$~\cite{cohen1960coefficient}. Over all trials, $\kappa = 0.72$ (87.7\% agreement);
by backbone, $\kappa$ is 0.65 on Stable Diffusion and 0.79 on Sana. These
values show sufficient agreement between the judgments of the two different
LLMs even at the level of individual trials. The above conclusions are
therefore unlikely to depend on the idiosyncrasies in the judgments of a
particular VQA model.

Fig.~\ref{fig:tab-qual-main} shows examples of images generated by the
compared methods and our method. In these examples, for the different visual
setups of a prompt with PP-attachment ambiguity, the compared methods fail to
depict the corresponding relations between objects, whereas our method
renders a distinct image that matches each visual setup.
Appendix~\ref{app:qualitative} shows more examples.

\subsection{Ablation Study}
\label{sec:ablation-pp}
We conduct an ablation study on the PP-attachment category of TAB by enabling
or disabling the two main components of CCG Diffusion for syntactic ambiguity
resolution: the Disambiguating Paraphrase and the Region Assignment Loss
(Section~\ref{sec:method-realization}). Table~\ref{tab:ablation-pp-effect}
shows the drop in the Yes rate, relative to the configuration with both
components, when either or both components are disabled.
On both backbones, disabling either component yields a lower Yes rate than
using both. On Stable Diffusion, the drop is statistically significant only
when both components are disabled; on Sana, it is significant only when the
Disambiguating Paraphrase or both components are disabled.

\begin{table}[t]
\caption{\textbf{Yes-rate drop from removing each component of CCG Diffusion on TAB PP.}
Yes rate (\%, judged by GPT-6 Astra) with each component or both disabled
minus the Yes rate with both components enabled (two-sided paired
permutation test over the 74 prompts, Holm-corrected over the three
comparisons per backbone). * $p<0.05$, ** $p<0.01$, *** $p<0.001$,
n.s.\ not significant.}
\label{tab:ablation-pp-effect}
\setlength{\tabcolsep}{4pt}
\centering\small
\begin{tabular}{l|c|c}
\hline
Removed component & Stable Diffusion & Sana\\
\hline
Disambiguating Paraphrase & $-$4.1\textsuperscript{n.s.} & $-$18.0\textsuperscript{***}\\
Region Assignment Loss & $-$2.0\textsuperscript{n.s.} & $-$0.9\textsuperscript{n.s.}\\
Both & $-$7.4\textsuperscript{**} & $-$17.7\textsuperscript{***}\\
\hline
\end{tabular}
\end{table}

As discussed for compositional text-to-image generation, one possible reason
why the drop differs in size between the backbones is that their text encoders
differ in representational capacity. The paraphrases for the two
interpretations consist of almost the same content words and differ only in
word order and syntactic structure (e.g., \prompt{It is the girl with a
telescope that the boy looks at.} and \prompt{With a telescope, the boy looks
at the girl.}). This difference may therefore be clearly reflected in the
conditioning of Sana, whose text encoder, Gemma, is an LLM that readily
captures syntactic structure, but only to a limited extent in the
conditioning of Stable Diffusion, whose text encoder, CLIP, poorly represents
word order and syntactic structure.

On the other hand, the drop from disabling the Region Assignment Loss is not
significant on either Stable Diffusion or Sana. On Stable Diffusion, whose
text encoder has weak representational capacity, the two components may each
make a small contribution and together support rendering the interpretations
as distinct images. On Sana, because Gemma has strong representational
capacity, the Disambiguating Paraphrase alone may suffice to render them as
distinct images, leaving almost no room for the Region Assignment Loss to
complement it.

\section{Conclusion}
\label{sec:conclusion}
In this paper, we introduce a CCG-based parser into text-to-image generation
models and propose CCG Diffusion, a training-free method that uses the
syntactic information in CCG derivations for prompt rewriting and attention
guidance. We apply it to two problems:
compositional text-to-image generation and syntactic ambiguity resolution.
Because the category of each word in CCG determines modification relations
and PP attachment, a single derivation provides the syntactic information
necessary for both problems. Experiments on benchmark datasets show that CCG Diffusion significantly
outperforms the compared baseline methods on both problems, except in some categories.
Experiments with two different backbones also show that CCG Diffusion is
effective without being tied to a specific model architecture.

Our work has the following limitations. First, CCG Diffusion has no mechanism
for controlling generation from sentences that involve depth relations,
actions, or multiple constraints, which are covered by the 3D-spatial,
non-spatial, and complex categories of T2I-CompBench++. Extending the guidance
to use depth cues and the predicate--argument structure of verbs is left for
future work. Second, the only type of syntactic ambiguity we address is PP
attachment. In other types of ambiguity, such as coordination scope
(\prompt{old men and women}), the interpretations can differ not in the category
of a single word but in how phrases are combined. These types therefore
require constraints on the derivation beyond fixing supertags, as well as
paraphrases and guidance designed for each ambiguity type. Third, the
gradient-based attention guidance increases the computational cost: generation
takes 2.2--2.4 times as long as vanilla generation (Appendix~\ref{app:cost}).

\appendices
\section{Implementation Details}
\label{app:implementation}
This appendix details the implementation and experimental settings of CCG
Diffusion.

\begin{table}[t]
\caption{\textbf{Generation settings and hyperparameters.}}
\label{tab:hyperparameters}
\setlength{\tabcolsep}{3pt}
\centering\small
\begin{tabular}{lcc}
\hline
 & Stable Diffusion & Sana\\
\hline
Resolution & \multicolumn{2}{c}{512$\times$512}\\
Denoising steps & 50 (PNDM) & 20\\
Classifier-free guidance & 7.5 & 4.5\\
Guided steps & first 25 & all 20\\
Extra updates at step 0 & 5 & 5\\
Attention maps & 5 layers (16$\times$16) & first 8 blocks\\
\hline
$(s,\delta)$, default (compositional) & \multicolumn{2}{c}{(20, 0.02)}\\
$(s,\delta)$, default (PP ambiguity) & \multicolumn{2}{c}{(10, 0.02)}\\
$(s,\delta)$, boosted (both tasks) & \multicolumn{2}{c}{(80, 0.06)}\\
$\gamma_{\mathrm{exist}}$, $\gamma_{\mathrm{corr}}$, $\gamma_{\mathrm{leak}}$, $\gamma_{\mathrm{poss}}$ & \multicolumn{2}{c}{0.1, 1, 0.3, 1}\\
$\lambda_{\mathrm{region}}$, $x_{\mathrm{b}}$, $\kappa$ & \multicolumn{2}{c}{0.25, 0.5, 0.08}\\
$\lambda_{\mathrm{spatial}}$, $m$, $r$ & \multicolumn{2}{c}{6, 0.25, 150/512}\\
$\lambda_{\mathrm{count}}$ (steps), $\theta$ & \multicolumn{2}{c}{0.1 (0--9), 0.3}\\
\hline
\end{tabular}
\end{table}

\begin{table}[t]
\caption{\textbf{Average generation time per image (s) on T2I-CompBench++.}}
\label{tab:cost}
\setlength{\tabcolsep}{3pt}
\centering\small
\begin{tabular}{lcc}
\hline
Method & Stable Diffusion & Sana\\
\hline
Vanilla & 1.77 & 0.80\\
Predicated Diffusion & 3.77 & 1.89\\
CCG Diffusion (ours) & 3.90 & 1.91\\
\hline
\end{tabular}
\end{table}

\begin{table*}[t]
\caption{\textbf{Score drop from removing two or more components on attribute binding in T2I-CompBench++.}
\checkmark\ marks the disabled components. Score of each configuration minus
the score with all three components enabled (two-sided paired permutation test
over the 300 prompts, Holm-corrected over the 12 comparisons per backbone).
* $p<0.05$, ** $p<0.01$, *** $p<0.001$.}
\label{tab:ablation-t2i}
\setlength{\tabcolsep}{4pt}
\centering\small
\begin{tabular*}{\textwidth}{@{\extracolsep{\fill}}ccc|ccc|ccc}
\hline
\multicolumn{3}{c|}{Removed component} & \multicolumn{3}{c|}{Stable Diffusion} & \multicolumn{3}{c}{Sana}\\
\begin{tabular}[b]{@{}c@{}}Selective\\Guidance Boost\end{tabular} & \begin{tabular}[b]{@{}c@{}}Positional\\Prefix\end{tabular} & \begin{tabular}[b]{@{}c@{}}Region\\Assignment Loss\end{tabular} & Color & Shape & Texture & Color & Shape & Texture\\
\hline
\checkmark & \checkmark & \checkmark & $-$0.070\textsuperscript{***} & $-$0.075\textsuperscript{***} & $-$0.068\textsuperscript{***} & $-$0.083\textsuperscript{***} & $-$0.069\textsuperscript{***} & $-$0.048\textsuperscript{***}\\
\checkmark & \checkmark & & $-$0.058\textsuperscript{***} & $-$0.069\textsuperscript{***} & $-$0.063\textsuperscript{***} & $-$0.079\textsuperscript{***} & $-$0.068\textsuperscript{***} & $-$0.047\textsuperscript{***}\\
\checkmark & & \checkmark & $-$0.037\textsuperscript{***} & $-$0.055\textsuperscript{***} & $-$0.054\textsuperscript{***} & $-$0.011\textsuperscript{**} & $-$0.010\textsuperscript{**} & $-$0.032\textsuperscript{***}\\
 & \checkmark & \checkmark & $-$0.046\textsuperscript{***} & $-$0.041\textsuperscript{***} & $-$0.017\textsuperscript{***} & $-$0.078\textsuperscript{***} & $-$0.047\textsuperscript{***} & $-$0.026\textsuperscript{***}\\
\hline
\end{tabular*}
\end{table*}

\subsection{Generation Settings and Hyperparameters}
Table~\ref{tab:hyperparameters} lists the hyperparameters used in our image
generation experiments. The size of a single update in
\eqref{eq:guidance} is determined by two values: a base scale $s$ and an
upper bound $\delta$. Let $T$ be the total number of denoising steps (50 for
Stable Diffusion and 20 for Sana), and let $j=T-t$ ($j=0,1,\ldots,T-1$) be the
index of step $t$ counted from the start of generation.
The step size at step $t$ is $\eta_t=s\sqrt{1-0.5\,j/(T-1)}$, which decreases
gradually from $s$ at $j=0$ to $s/\sqrt{2}$ at $j=T-1$. To prevent a single
update from changing the latent too much, if the L2 norm of the update
$\eta_t\nabla_x\mathcal{L}$ exceeds $\delta$ times the L2 norm of the latent
$x$, we scale the update down to an L2 norm of $\delta\|x\|_2$ without changing
its direction. The guidance strength in
Section~\ref{sec:method-guidance-strength} refers to this pair $(s,\delta)$.
For inputs to which the Selective Guidance Boost applies, we use the boosted
pair instead of the default pair.

For the Count Loss (Section~\ref{sec:method-count}), let $\widetilde{A}_n$ be
the $16\times16$ attention map of noun $n$ smoothed with a $3\times3$ Gaussian
kernel (standard deviation 0.5); $p_n$ is $\widetilde{A}_n$ normalized to sum
to 1. The target $q_n$ is an equally weighted mixture of $k_n$ Gaussian peaks.
We choose their centers greedily from the most strongly attended pixels
(min-max normalized value at least $\theta$) so that they are at least
$\min(16/k_n,4)$ pixels apart. Each peak is confined to the Voronoi cell of its
center, and its width is set by half the distance to the nearest other center
(at most 2 pixels), so that the peaks do not overlap. The weight $g_n$ in
\eqref{eq:count-loss} is $\min\bigl(1,(77/I)\sum_i\widetilde{A}_n[i]\bigr)$,
which falls below 1 when the attention of the noun is weak, and no gradient
flows through it. We add the Count Loss only at steps where the attention of
every counted noun is neither vanishing nor flat, i.e., where both the mean and
the range of $\widetilde{A}_n$ are at least $10^{-4}$.

\subsection{Conditions for Applying the Components}
\label{app:conditions}
For compositional text-to-image generation, we apply each component of CCG
Diffusion only to input prompts for which the corresponding structure can be
read off from the derivation.
We apply the Selective Guidance Boost to inputs in which a modifier bound to a
noun belongs to a fixed vocabulary of color, shape, size, material, and surface
words. However, for inputs that contain cardinal numbers, 2D relation phrases,
or depth words, we use the standard guidance strength.
We apply the Positional Prefix and the Region Assignment Loss to input prompts
in which two non-overlapping noun phrases, each with a color, texture, shape,
or size modifier, are joined by \prompt{and}, provided that neither the Count
Loss nor the Spatial Relation Loss applies to the prompt.

\subsection{Hardware}
\label{app:hardware}
We ran the compositional text-to-image generation experiments on an NVIDIA
DGX Station A100 with an AMD EPYC 7742 CPU (64 cores) and four NVIDIA A100
SXM4 GPUs (80~GB each). For the syntactic ambiguity resolution experiments,
we generated images on a machine with an Intel Core i9-14900KF CPU (24 cores)
and one NVIDIA GeForce RTX 4090 GPU (24~GB), and ran the automatic evaluation
with Qwen3.8-Flash-Next on two NVIDIA DGX Spark systems. Each DGX Spark has a
20-core Arm CPU (ten Cortex-X925 and ten Cortex-A725 cores) and an NVIDIA GB10
Grace Blackwell Superchip. We connected the two systems directly with a QSFP
cable through their ConnectX-7 interfaces and distributed the model across them
for inference.

\subsection{Computational Cost}
\label{app:cost}
Table~\ref{tab:cost} shows the generation time of the vanilla backbones,
Predicated Diffusion, and our method, measured over the whole pipeline from
receiving a prompt to obtaining the image. We randomly selected 30 prompts from
each of the eight subcategories of T2I-CompBench++ (240 in total) and
generated one image per prompt on a single A100 GPU (80~GB) of the DGX Station
A100 described in Appendix~\ref{app:hardware}.

Our method takes about 2.2 times (Stable Diffusion) and about 2.4 times (Sana)
as long as the vanilla backbone, and about 4\% (Stable Diffusion) and about 1\% (Sana)
longer than its base method, Predicated Diffusion. Most of the difference from
the vanilla backbone comes from computing the gradient of the Predicated
Diffusion loss and updating the latent, which also take a similar amount of
time in Predicated Diffusion. As for the difference from Predicated Diffusion,
the CCG parser used in this work (Hol-CCG) is more expensive than spaCy's
dependency parser (parsing takes about 25--45~ms longer). However, parsing
accounts for only about 1--3\% of the runtime of the whole image generation
pipeline, so its effect on the total generation time is limited.

\section{Full Results of the Ablation Study on Compositional Text-to-Image Generation}
\label{app:ablation-full}
Table~\ref{tab:ablation-t2i} shows the change in score when two or more
components of CCG Diffusion are disabled in the ablation study of
Section~\ref{sec:ablation-compositional}.

\section{Additional Qualitative Examples}
\label{app:qualitative}
\begin{figure*}[!tp]
\centering
\includegraphics[height=8.5in]{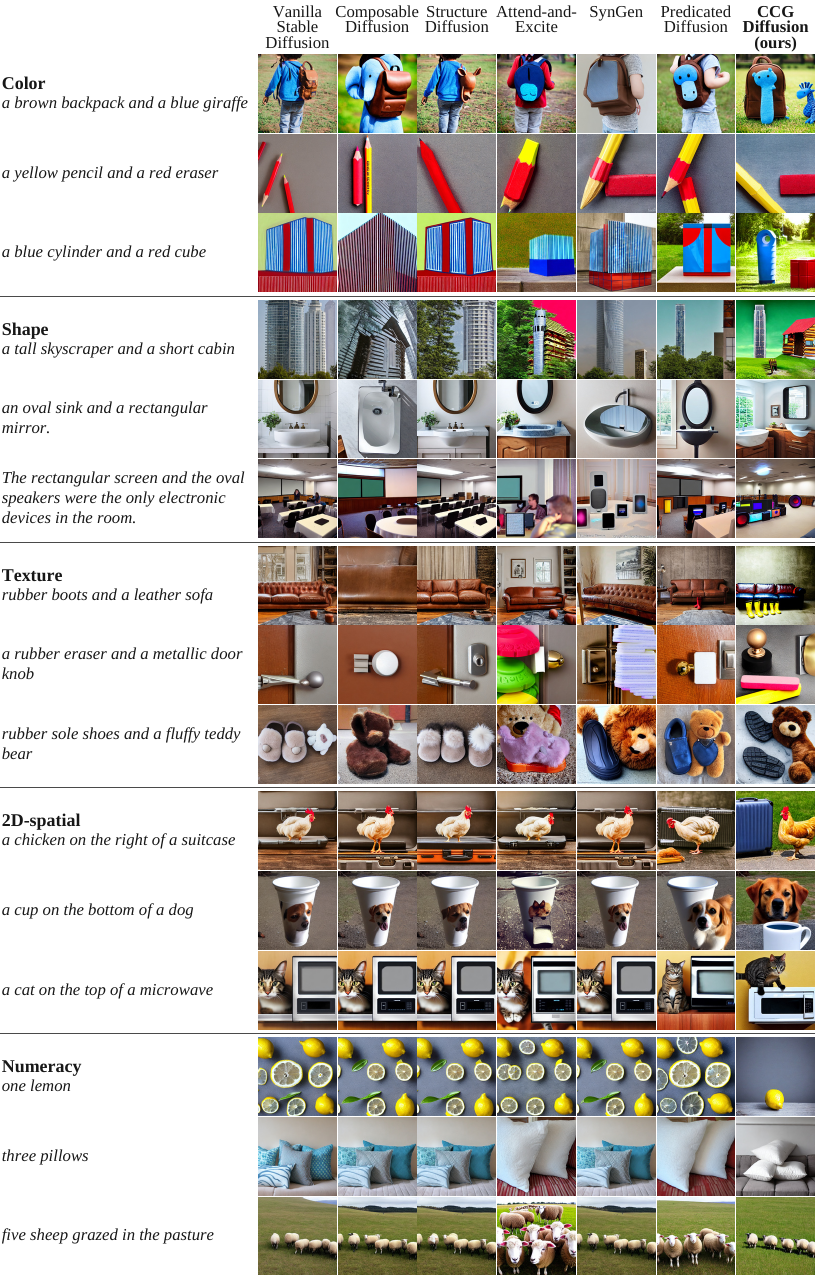}
\caption{\textbf{Additional Stable Diffusion examples: Color, Shape, Texture, 2D-spatial, and Numeracy.}
}
\label{fig:t2i-app-sd14}
\end{figure*}

\begin{figure*}[!tp]
\centering
\includegraphics[height=8.5in]{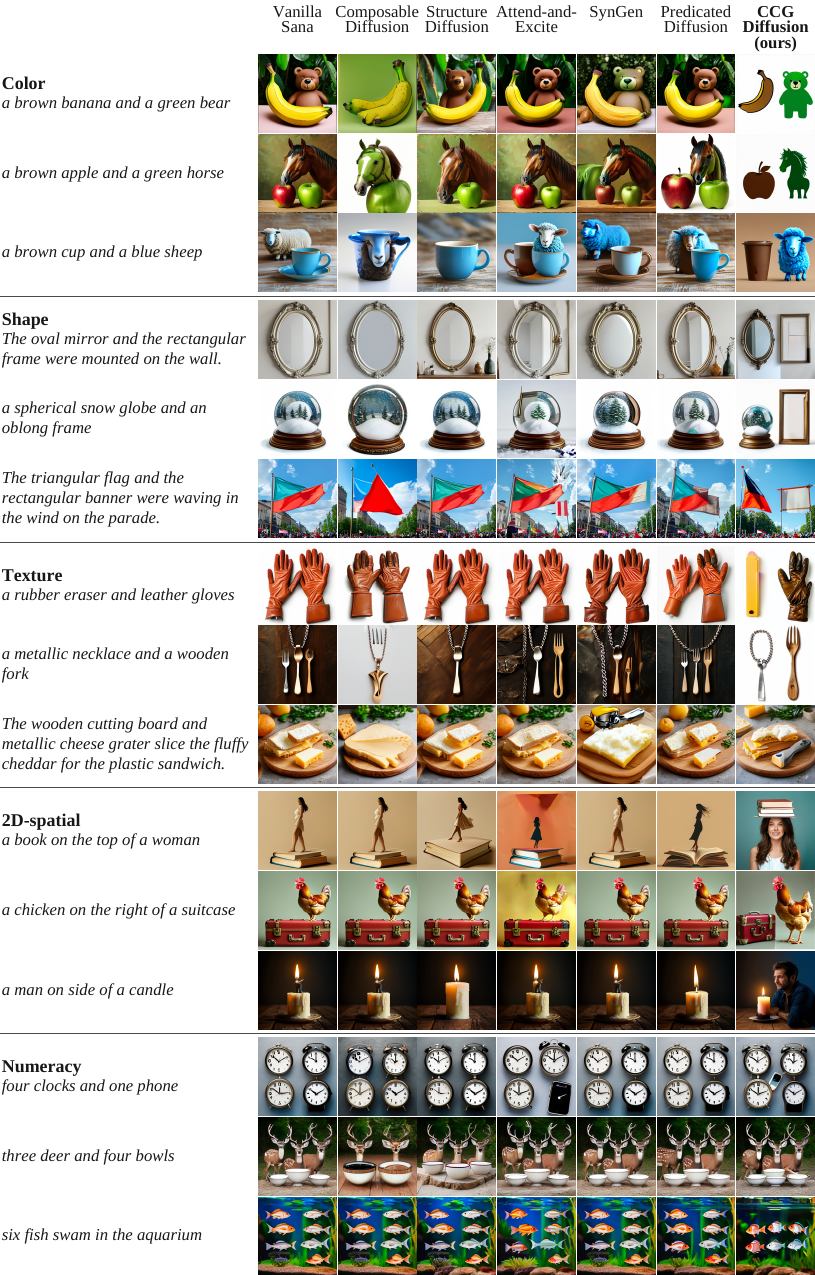}
\caption{\textbf{Additional Sana examples: Color, Shape, Texture, 2D-spatial, and Numeracy.}
}
\label{fig:t2i-app-sana}
\end{figure*}

\begin{figure*}[!tp]
\centering
\includegraphics[width=\textwidth,height=0.86\textheight,keepaspectratio]{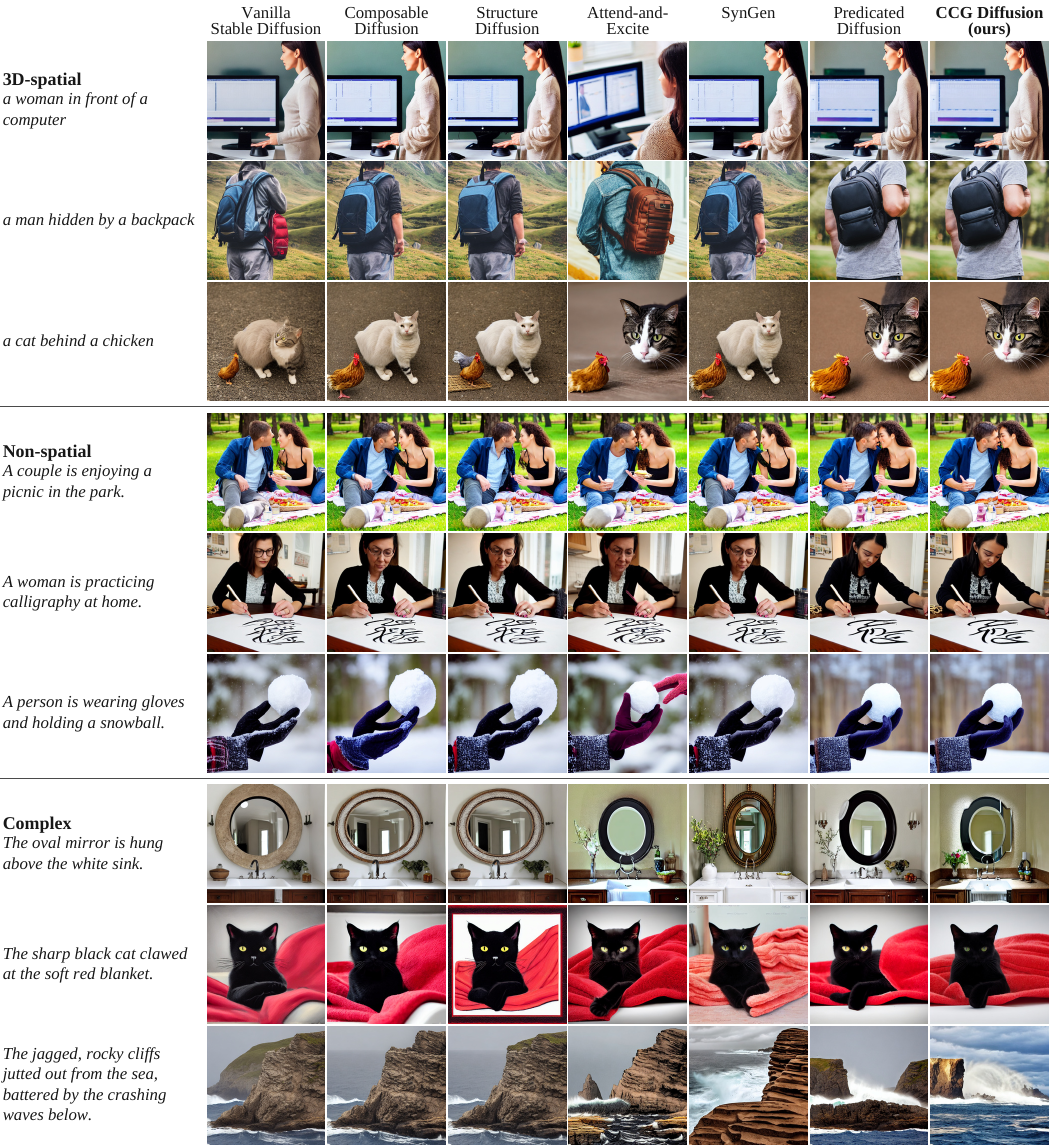}
\caption{\textbf{Additional Stable Diffusion examples: 3D-spatial, Non-spatial, and Complex.}
}
\label{fig:t2i-app-sd14-3d-ns-cx}
\end{figure*}

\begin{figure*}[!tp]
\centering
\includegraphics[width=\textwidth,height=0.86\textheight,keepaspectratio]{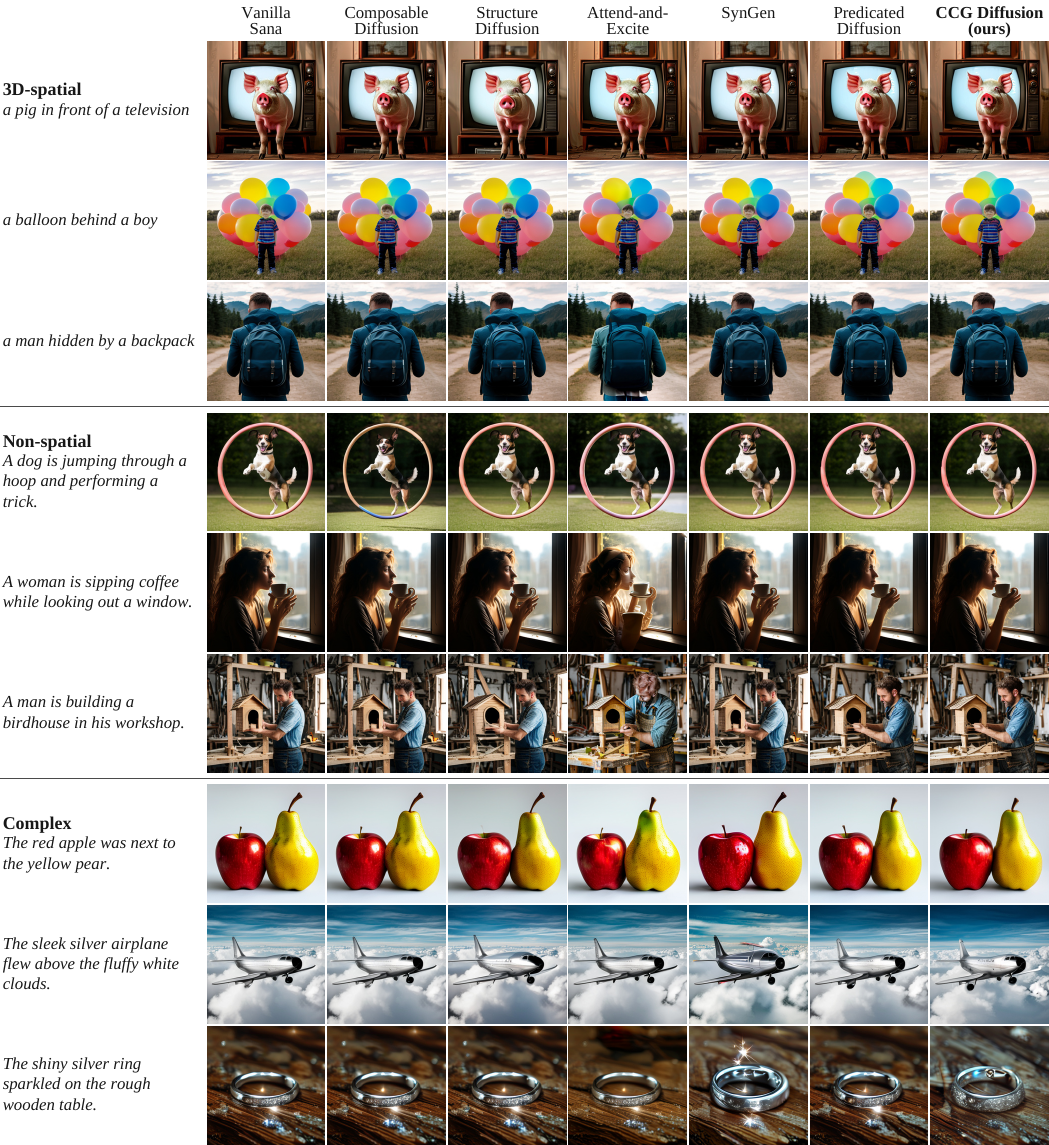}
\caption{\textbf{Additional Sana examples: 3D-spatial, Non-spatial, and Complex.}
}
\label{fig:t2i-app-sana-3d-ns-cx}
\end{figure*}

Figs.~\ref{fig:t2i-app-sd14} and~\ref{fig:t2i-app-sana} show additional
examples on Stable Diffusion and Sana in the color, shape, texture,
2D-spatial, and numeracy categories of compositional text-to-image generation.
Figs.~\ref{fig:t2i-app-sd14-3d-ns-cx} and~\ref{fig:t2i-app-sana-3d-ns-cx} show
examples in the 3D-spatial, non-spatial, and complex categories.

\begin{figure*}[!tp]
\centering
\includegraphics[width=\textwidth]{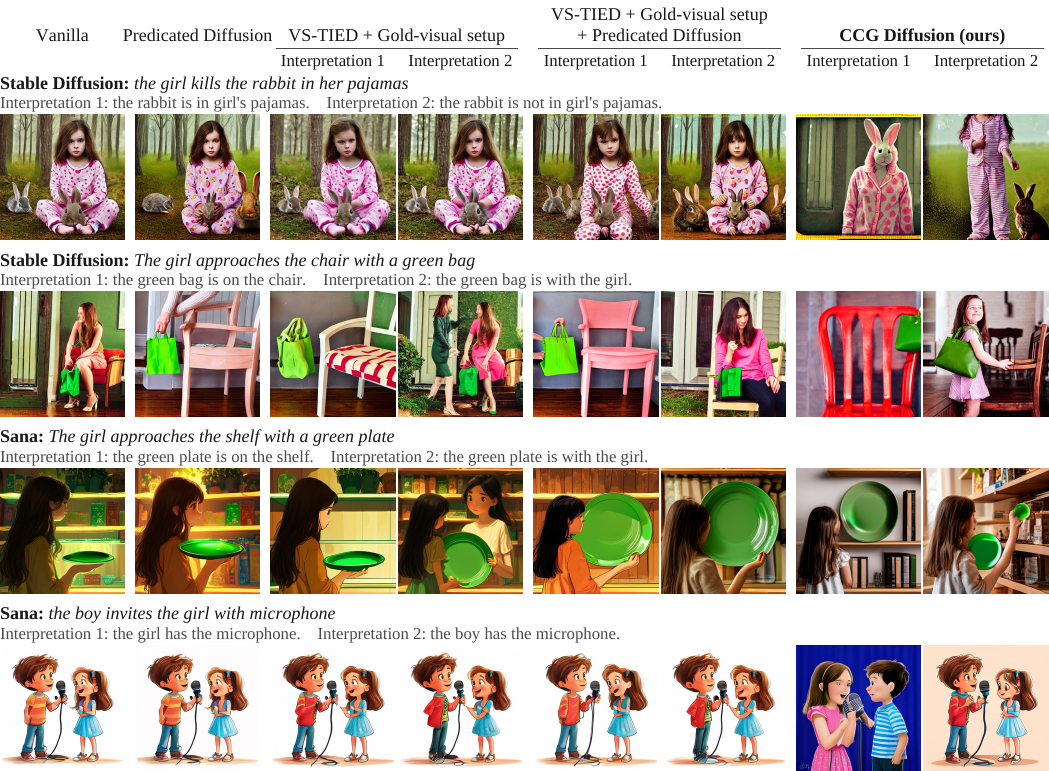}
\caption{\textbf{Additional examples of generated images for
the TAB PP-attachment category on Stable Diffusion and Sana.}
}
\label{fig:tab-qual-app}
\end{figure*}

Fig.~\ref{fig:tab-qual-app} shows additional examples on Stable Diffusion and
Sana for PP ambiguity resolution.

\clearpage
\makeatletter\global\@colroom\@colht\global\vsize\@colht\makeatother
\section*{Acknowledgment}
In this work, we used two AI coding agents: Codex (OpenAI; GPT-5 and GPT-6
series models)~\cite{codex} and Claude Code (Anthropic; Claude Opus 5 and
Claude Fable 5 series models)~\cite{claudecode}.

We used Codex mainly to implement the experiments. Following our
specifications, Codex wrote new source code, debugged and refactored code,
ran the experiments, aggregated the experiment logs, and created the scripts
that generate the figures and tables.

We used Claude Code mainly to write this manuscript. For every section of this
paper, we specified the content to be written, had Claude Code write a draft,
and then checked and revised the draft. We also used Claude Code for English
proofreading, for \LaTeX{} formatting and table typesetting, and for
suggesting candidate references.

Both tools also served as discussion partners when we designed the method and
experiments and when we interpreted the results, but all research ideas,
design decisions, and interpretations of the results are our own. We verified
AI-generated code through tests and comparison with known reference results,
and we checked every number reported in this paper against the raw experiment
logs. We take full responsibility for the content of this paper.

\bibliographystyle{IEEEtran}
\bibliography{refs}

\begin{IEEEbiography}[{\includegraphics[width=1in,height=1.25in,clip,keepaspectratio]{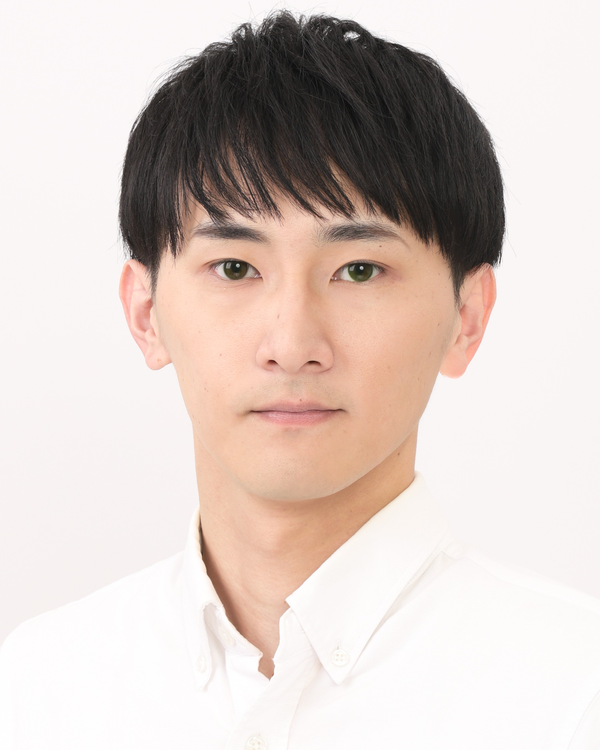}}]{Ryosuke Yamaki}
received the M.E. degree from Ritsumeikan University, Japan, in 2023, where
he is currently pursuing the Ph.D. degree with the Graduate School of
Information Science and Engineering. His research interests include natural
language processing, syntactic parsing, and text-to-image generation. He is
a Student Member of the Association for Natural Language Processing.
\end{IEEEbiography}

\begin{IEEEbiography}[{\includegraphics[width=1in,height=1.25in,clip,keepaspectratio]{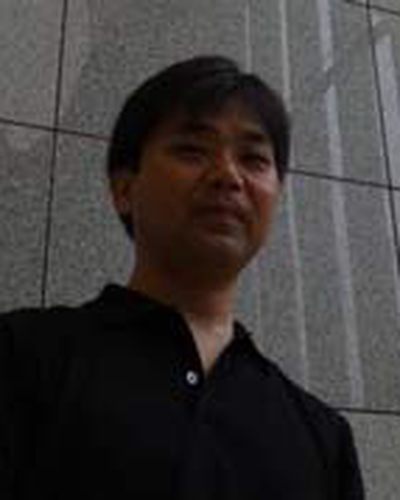}}]{Daichi Mochihashi}
received the B.S. degree from The University of Tokyo, in 1998, and the
Ph.D. degree from Nara Institute of Science and Technology, in 2005. He is
currently a Professor with The Institute of Statistical Mathematics and
jointly with the National Institute for Japanese Language and Linguistics,
Japan. His research interests include natural language processing, machine
learning, and Bayesian statistics.
\end{IEEEbiography}

\begin{IEEEbiography}[{\includegraphics[width=1in,height=1.25in,clip,keepaspectratio]{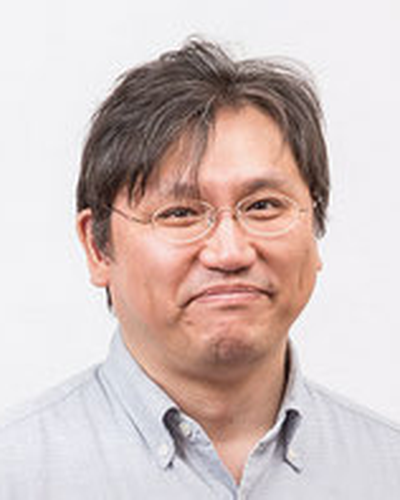}}]{Nobutaka Shimada}
(Member, IEEE) received the M.E. and Ph.D. degrees from Osaka University, in
1994 and 1997, respectively. From 1997 to 2003, he was a Research Associate
at the Department of Mechanical Engineering for Computer-Controlled
Machinery, Osaka University. From 2004 to 2012, he was an Associate
Professor at the Department of Human and Computer Intelligence, Ritsumeikan
University. From 2007 to 2008, he was a Visiting Associate Professor at the
Robotics Institute, Carnegie Mellon University. Since 2012, he has been a
Professor at the Department of Information and Engineering, Ritsumeikan
University. His research interests include computer vision, robot vision,
and human--computer interaction. He is a member of IEEE CS, IEICE, IPSJ,
RSJ, JSAI, and HIS.
\end{IEEEbiography}

\begin{IEEEbiography}[{\includegraphics[width=1in,height=1.25in,clip,keepaspectratio]{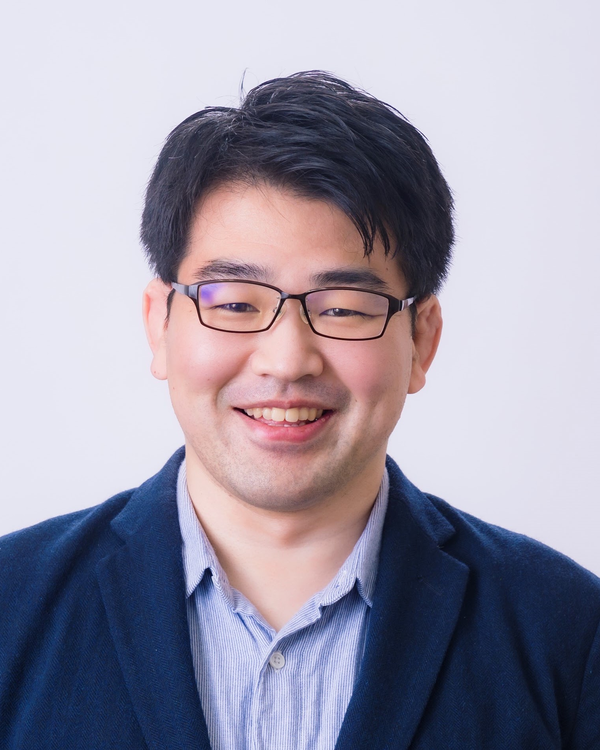}}]{Seitaro Shinagawa}
graduated from the Faculty of Engineering at Tohoku University in 2013.
He completed his Master’s degree in Information
Science at the same university in 2015, and his
doctoral studies at the Nara Institute of Science and
Technology (NAIST) in 2020, earning a Doctor of
Engineering. After serving as an Assistant Professor at NAIST, he joined SB Intuitions Corp. and
became an Affiliate Assistant Professor at NAIST
in 2024. He is engaged in the research and development of multimodal foundation models.
\end{IEEEbiography}

\begin{IEEEbiography}[{\includegraphics[width=1in,height=1.25in,clip,keepaspectratio]{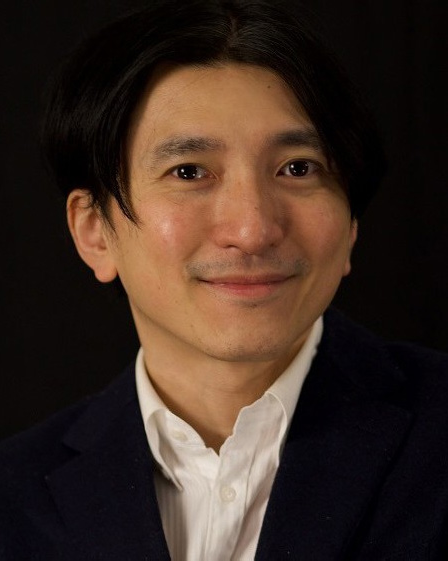}}]{Tadahiro Taniguchi}
(Member, IEEE) received the M.E. and Ph.D. degrees from Kyoto University, in
2003 and 2006, respectively. From 2005 to 2008, he was a Japan Society for
the Promotion of Science Research Fellow with Kyoto University. From 2008 to
2010, he was an Assistant Professor with the Department of Human and
Computer Intelligence, Ritsumeikan University. From 2010 to 2017, he was an
Associate Professor with the Department of Human and Computer Intelligence,
Ritsumeikan University. From 2015 to 2016, he was a Visiting Associate
Professor with the Department of Electrical and Electronic Engineering,
Imperial College London. From 2017 to 2024, he was a Professor with the
Department of Information and Engineering, Ritsumeikan University. He is
currently a Professor with the Graduate School of Informatics, Kyoto
University, and an Affiliate Professor with the Research Organization of
Science and Technology, Ritsumeikan University. He has been engaged in
research on machine learning, emergent systems, intelligent vehicles, and
symbol emergence in robotics.
\end{IEEEbiography}

\EOD

\end{document}